\documentclass[11pt,reqno]{article}
\usepackage{fullpage,graphicx,xcolor,natbib}
\usepackage[colorlinks,citecolor=bbluegray,linkcolor=ddarkbrown,urlcolor=blue,breaklinks]{hyperref}

\usepackage{amssymb}
\usepackage{amsthm} 
\usepackage{amsmath} 

\usepackage{mathtools}
\mathtoolsset{showonlyrefs}
\usepackage{xparse}

\usepackage{color}

\usepackage{authblk}

\usepackage[utf8]{inputenc}
\usepackage[T1]{fontenc}
\usepackage{newtxtext}
\usepackage{booktabs}       
\usepackage{microtype}      
\usepackage{natbib}         

\usepackage{etoolbox}
\usepackage[enable]{easy-todo}

\usepackage{graphicx}
\usepackage{multirow}

\usepackage{subfig}
\usepackage{enumitem}

\usepackage{algorithmic,algorithm}

\let \oldsection \section
\renewcommand{\section}{\vspace{3ex plus 1ex}\oldsection}

\definecolor{ddarkbrown}{rgb}{0.5,0.2,0.05} \definecolor{bbluegray}{rgb}{0.05,0,0.5}

\def\xx{{\boldsymbol x}}

\newcommand{\BEAS}{\begin{eqnarray*}}
\newcommand{\EEAS}{\end{eqnarray*}}
\newcommand{\BEA}{\begin{eqnarray}}
\newcommand{\EEA}{\end{eqnarray}}
\newcommand{\BEQ}{\begin{equation}}
\newcommand{\EEQ}{\end{equation}}
\newcommand{\BIT}{\begin{itemize}}
\newcommand{\EIT}{\end{itemize}}
\newcommand{\BNUM}{\begin{enumerate}}
\newcommand{\ENUM}{\end{enumerate}}

\newcommand{\BA}{\begin{array}}
\newcommand{\EA}{\end{array}}

\newcommand{\argmin}{\mathop{\rm argmin}}

\newcommand{\norm}[1]{\left\lVert#1\right\rVert}

\title{Trace-Norm Adversarial Examples}

\author[1]{Ehsan Kazemi} 

\author[2,3]{Thomas Kerdreux} 
\author[1]{Liqiang Wang}

\affil[1]{Department of Computer Science, University of Central Florida, Florida, USA.}
\affil[2]{D.I., UMR 8548, \'Ecole Normale Sup\'erieure, Paris, France.}
\affil[3]{INRIA, Paris, France}

\begin{document}

\maketitle

\begin{abstract}
White box adversarial perturbations are sought via iterative optimization algorithms most often minimizing an adversarial loss on a $\ell_p$ neighborhood of the original image, the so-called distortion set. Constraining the adversarial search with different norms results in disparately structured adversarial examples. Here we explore several distortion sets with structure-enhancing algorithms. These new structures for adversarial examples, yet pervasive in optimization, are for instance a challenge for adversarial theoretical certification which again provides only $\ell_p$ certificates. Because adversarial robustness is still an empirical field, defense mechanisms should also reasonably be evaluated against differently structured attacks. Besides, these structured adversarial perturbations may allow for larger distortions size than their $\ell_p$ counter-part while remaining imperceptible or perceptible as natural slight distortions of the image. Finally, they allow some control on the \textit{generation} of the adversarial perturbation, like (localized) bluriness.
\end{abstract}

\section{Introduction}\label{sec:introduction}
Adversarial examples are inputs to machine learning classifiers designed to cause the model to misclassify. These are searched in the vicinity of some dataset samples, typically in their norm-ball neighborhoods, the so-called \textit{distortion set}. When replacing every test set samples with their associated sought for adversarial examples, the accuracy of classically trained classifiers quickly drops to zero as a function of the considered norm-ball radius. This lack of robustness challenges the security of some real-world systems as well as questions the neural classifiers generalizing properties \citep{schmidt2018adversarially,stutz2019disentangling}.

This has hence sired a series of works proposing attacks or defenses methods (both practical or theoretical). Most of the attack and defense mechanisms considered $\ell_p$ neighborhoods. While some recent papers \citep{xu2018structured,wong2019wasserstein} pointed out the benefits of other families of distortions sets, as well as others, outlined the inherent limitations of the $\ell_p$ balls \citep{sharif2018suitability,sen2019should}, many classical norm families remained mostly unexplored in the adversarial setting.

\begin{figure}[h!]
    \centering
    \subfloat[FGSM]{{\includegraphics[scale=0.3]{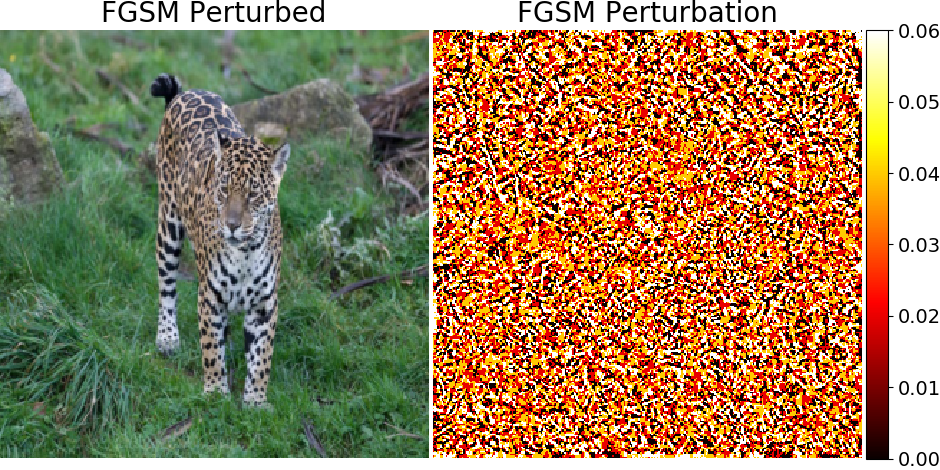}}}\hfill
    \centering
    \subfloat[With nuclear ball]{{\includegraphics[scale=0.3]{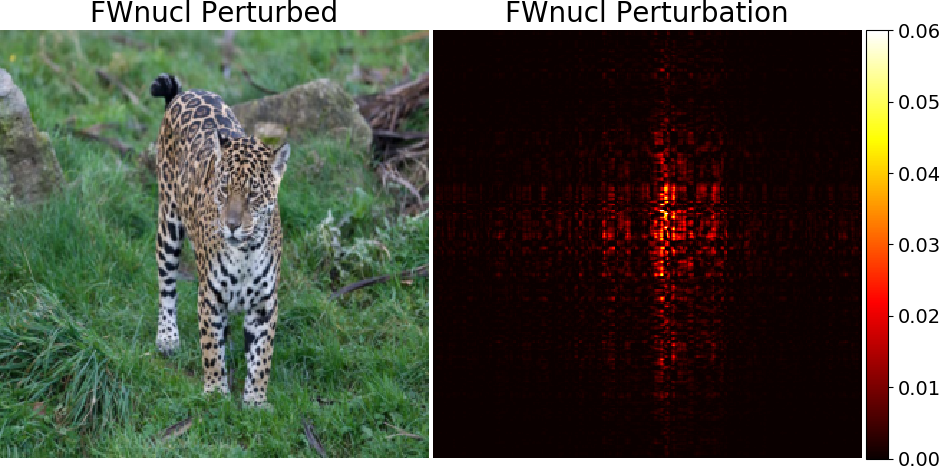}}}
    \caption{The images correspond to two types of targeted attacks. Projected Gradient Descent (PGD) solve \eqref{eq:opt_problem_adversarial} constrained by a $\ell_{\infty}$ ball while FWnucl solves \eqref{eq:opt_problem_adversarial} constrained with a nuclear ball. The type of adversarial perturbations differs significantly in structure.}
    \label{fig:example_diff_distortion}
\end{figure}

In the white-box framework,  the attacker has access to backward passes on the model. Adversarial examples are then iterates of optimization algorithms that seek to minimize constrained adversarial losses. Although norms are equivalent in the image finite-dimensional space, the type of norm-ball influences the structure of the optimization algorithm iterates as well as the (local) minima to which they seek to converge. In such an empirical field, it is hence essential to explore the effect of particular structures in adversarial perturbation besides $\ell_p$ balls, see Figure \ref{fig:example_diff_distortion} for an example.

For instance, it may perfectly be that some empirical defense mechanisms leverage on the lack of a certain pattern in the adversarial perturbation. Providing a catalog of many structured attacks would then at least warn us against such a possibility.

The radius of the convex balls is often taken small enough to ensure that the adversarial perturbations are imperceptible. This imperceptibility requirement is pervasive in the literature, although it is far from being the only regime for adversarial examples \citep[\S 2]{gilmer2018motivating} leading to security issues.
Adversarial examples also challenge the generalization properties of classifiers. However, the imperceptibility of the distortion does not play a special role in that respect, especially when the ideal level of perturbation is arguably the one that guarantees that the labeled content of the original image is preserved (for a human observer) in the adversarially perturbed image.
Actually, the imperceptible deformation regime of non-robust classifiers has arguably received much attention because it \textit{highlights the limitation of the analogy between human perception and the processing done by machine learning systems} \citep{gilmer2018motivating}.
Not being able to overcome this issue would then challenge the belief that current neural networks are possible models of some biological brains.

Here, we do not limit ourselves to the imperceptible regime of perturbation. Instead, we explore adversarial examples' structure leading to possibly perceptible deformations that would yet be considered as \textit{non-suspicious} alteration of the image. For instance, the nuclear ball, which is the convex relaxation of matrix rank, qualitatively leads to blurred versions of the initial image.
This blurring effect could be easily localized to specific semantic areas in the image simply by considering the group-nuclear ball distortion set, where the groups are defined accordingly.

For the sake of simplicity, we focus on un-targeted adversarial examples via the classical optimization approach with respect to $x$
\begin{eqnarray}\label{eq:opt_problem_adversarial}
\mbox{minimize} & L(f(x), t)\\
\mbox{subject to} &  ~ ||x-x^{ori}|| \leq \epsilon
\end{eqnarray}
where $L$ is an adversarial loss, $f$ the neural classifier and $t$ a label different from the label of the original image $x_{ori}$.

\paragraph{Related Work.}
Several works question the reason for considering $\ell_p$ neighborhood as distortion sets and propose other models and methods. For instance, \cite{sharif2018suitability} suggest that $\ell_p$ norms are neither the right metric for perceptually nor even content-preserving adversarial examples. \cite{sen2019should} show on a behavioral study that that $\ell_p$ norms and some other metrics do not fit with human perception.

Others consider adversarial perturbations beyond the $\ell_p$ distortion sets. \cite{engstrom2017rotation} show that simple rotation and translation can be efficient adversarial methods.
\cite{xu2018structured} consider group-lasso distortion sets to latter interpret some adversarial example properties, solving the optimization problem via ADMM.
\cite{liu2018beyond} generate adversarial examples based on the geometry and physical rendering of the image. They notably suggest that \textit{large pixel perturbations can be realistic if the perturbation is conducted in the physical parameter space (e.g., lighting)}.
\citep{wong2019wasserstein} recently argue that robustness to Wasserstein perturbations of the original image is essentially an invariance that should typically exist in classifiers. 

Some methods solve the adversarial optimization problem on subspaces, which might lead to specifically structured adversarial examples. While random subspace \citep{Subspace19} does not induce specific perturbation structure, projection on low-frequency domain \citep{guo2018low} or onto the subspace generated by the top few singular vectors of the image \citep[\S 3.4.]{yang2019me} do induce structured adversarial examples. These approaches aim at reducing the search space of adversarial perturbation for computational reasons.

Finally, one can consider the problem of adversarial attack generation as an image processing task. A recent trend to various types of such algorithms, like in conditional or unconditional GAN, style transfer algorithms, or image translation algorithms, has been to provide more control for the user of the modified or generated images \citep{reed2016learning,gatys2017controlling,risser2017stable,lu2017decoder}. Providing a little more control to the attacker on the generated type of adversarial perturbations stands in that line of works.

\paragraph{Contribution.} We study some families of structured norms in the adversarial example setting. This is a pretense to more generally motivate the relevance of structured attacks (\textit{i.e.} besides the $\ell_p$ distortion set), that are largely unexplored.
For instance, it challenges the theoretical certification approaches that are stated in terms of $\ell_p$ neighborhoods. It is also a versatile approach to produce specific modification of the adversarial images, like (local) bluriness.

\paragraph{Outline.}
In Section \ref{sec:structured_distortion_sets} we motivate simple families of norm resulting in structured adversarial perturbations which can then be leveraged for crafting specific adversarial attacks. In Section \ref{sec:FW_for_adversarial_examples} we use the conditional gradient algorithms designed to efficiently solve \eqref{eq:opt_problem_adversarial} under the various considered structured norms and we then report some numerical experiments in Section \ref{sec:numerical_experiments}.

\section{Structured Distortion Sets}\label{sec:structured_distortion_sets}

Here we detail some simple structured families of norms that, to the best of our knowledge, have not yet been explored in the context of adversarial attacks. When solving \eqref{eq:opt_problem_adversarial}, these lead to specific structures in the outputs, giving some control, for instance to a potential attacker, on the way the adversarial perturbation alters the test sample images. In addition, these perturbations can be further adapted to the attacked image by solving the adversarial problem \eqref{eq:opt_problem_adversarial} with a group-norm distortion sets, where the groups are adapted to the image.

In Section \ref{sec:FW_for_adversarial_examples}, we seek to solve the optimization problem \eqref{eq:opt_problem_adversarial} with conditional gradient algorithms. 
When applicable, this ensures that the early algorithms' iterates have a specific structure.
Indeed, each iteration of these algorithms requires to solve a Linear Minimization Oracle (LMO). For a direction $d$ and a convex set $\mathcal{C}$, it is defined as
\BEQ\label{eq:general_definition_LMO}
\text{LMO}_{\mathcal{C}}(d) \in\underset{v\in\mathcal{C}}{\argmin } ~ d^T v.
\EEQ
The iterates of conditional gradient algorithms are then constructed as (sparse) convex combination of such solutions. These can always be chosen as extreme points of $\mathcal{C}$. Hence, the specific structure of the solutions of the LMO is then passed on to the early optimization iterates. In Section \ref{ssec:group_constraints}, we review how to leverage on weighted group norms in order to localize the low-rank perturbations.

\subsection{Low-rank perturbation}\label{ssec:low_rank_perturbation}
We write $||\cdot||_{S1}$ for the nuclear norm, the sum of the matrix singular value, a.k.a. the trace norm or the $1$-Schatten norm. The nuclear norm has been classically used to promote low-rank solutions of convex optimization problems \citep{fazel2001rank,candes2009exact} such as matrix completion. 
Here, we propose to simply consider nuclear balls as distortion sets when searching for adversarial examples in problem \eqref{eq:opt_problem_adversarial}. We later explore the various benefit of using this structure.
To our knowledge, the low-rank structure is used in different aspects of some defense techniques but not for adversarial attacks. 
As an empirical defense mechanism, \cite{langeberg2019effect} add a penalization in the training loss that promotes the low-rank structure of the convolutional layers filters. \cite{yang2019me} notably propose a pre-processing of the classifier outputs, which randomly removes some input pixels and further reconstructs it via matrix completion. We hence seek to solve the following optimization problem

\begin{equation}\label{eq:low_rank_adversarial}\tag{lr-OPT}
\underset{||x-x^{ori}||_{S1} \leq \epsilon}{\argmin } L(f(x), t).
\end{equation}
This is just one example among the family of $p$-Schatten norms $||\cdot||_{Sp}$, the $p$-norm of the singular value vector. These lead to differently structured adversarial examples and may be interesting to defeat certification approaches that are stated in terms of $\ell_p$ neighborhoods.
The LMO \eqref{eq:general_definition_LMO} for a nuclear ball of radius $\rho$ is given as
\BEQ\label{eq:LMO_nuclear}
\text{LMO}_{||\cdot||_{S1}\leq \rho}(M) \triangleq \rho~U_1 V_1^T,
\EEQ
when the singular decomposition of the matrix $M$ is $U S V^T$. The computation of the LMO involves $U_1$ and $V_1$, i.e., the right and left singular vectors associated to the largest singular value $\rho$. Others optimization approaches like PGD, would require the knowledge of the full SVD at each iteration. Note that for $q$-Schatten norm (with $q>1$) the LMO is also explicit and involve the full singular decomposition (see \citep[Lemma 7]{garber2015faster} for instance).

\begin{figure}
    \centering
    \subfloat[Original]{\includegraphics[width=0.19\linewidth]{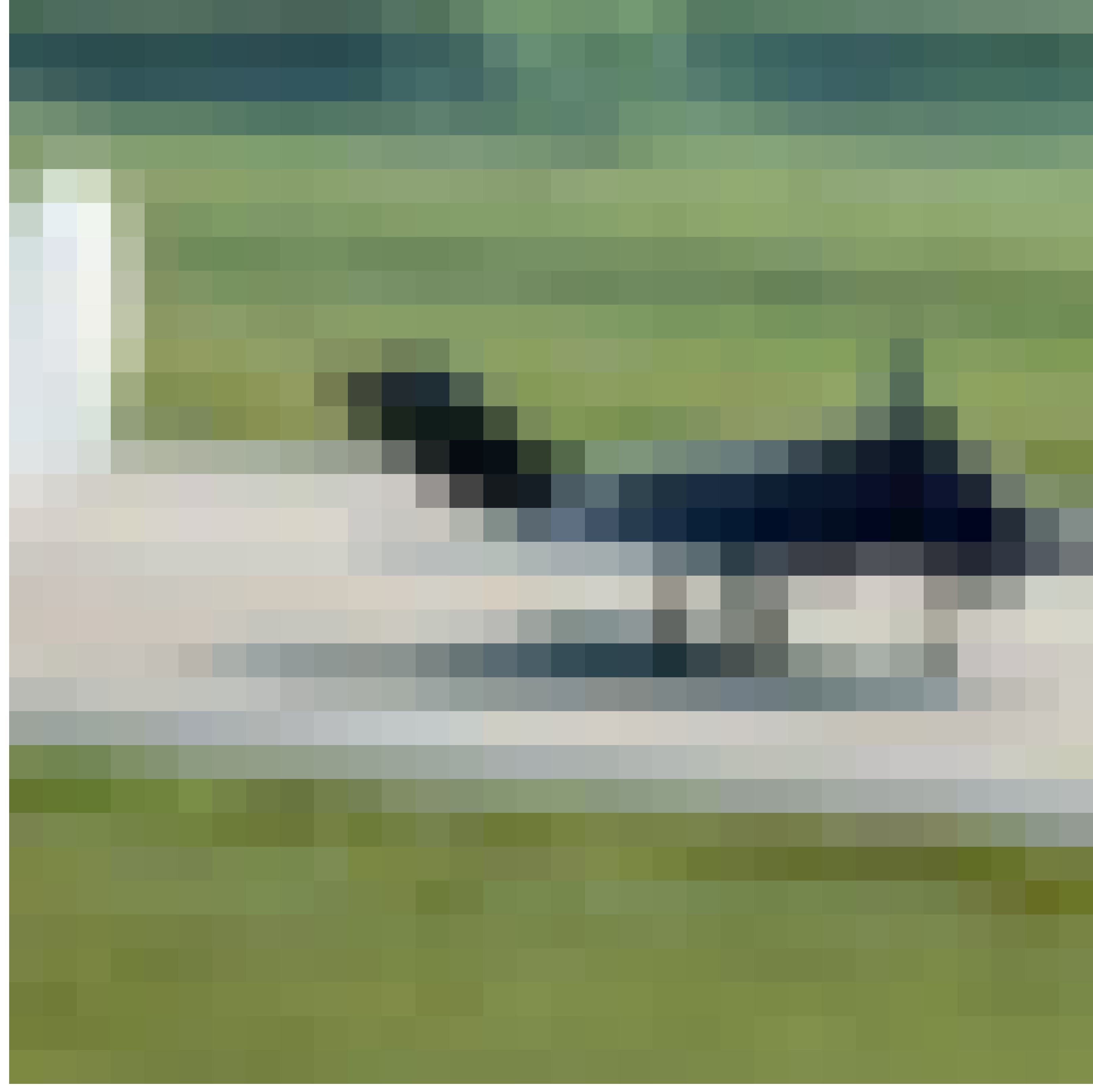}}
    \subfloat[$\epsilon_{S1}=5$]{\includegraphics[width=0.19\linewidth]{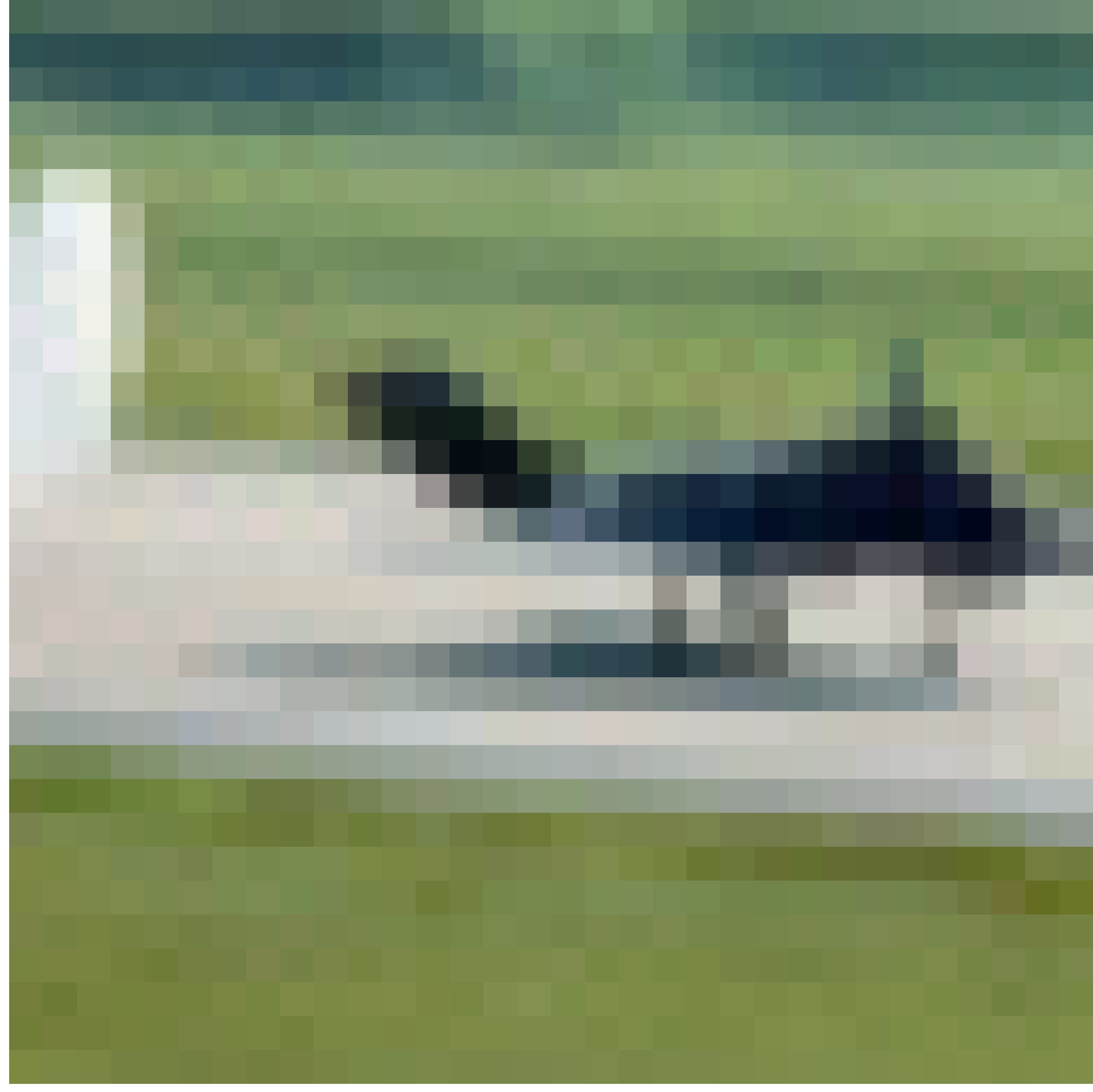}}
    \subfloat[$\epsilon_{S1}=10$]{\includegraphics[width=0.19\linewidth]{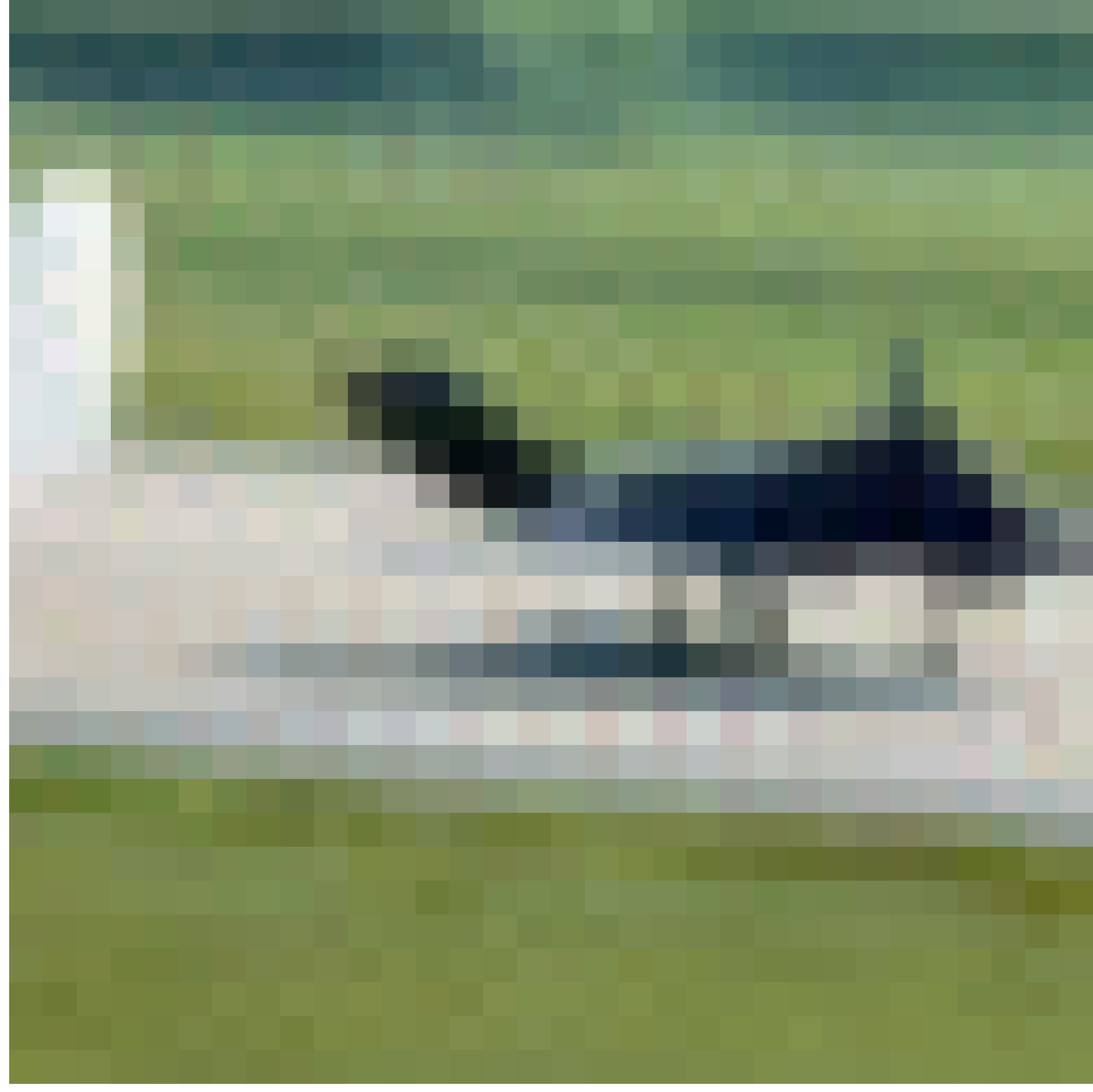}}
    \subfloat[$\epsilon_{S1}=20$]{\includegraphics[width=0.19\linewidth]{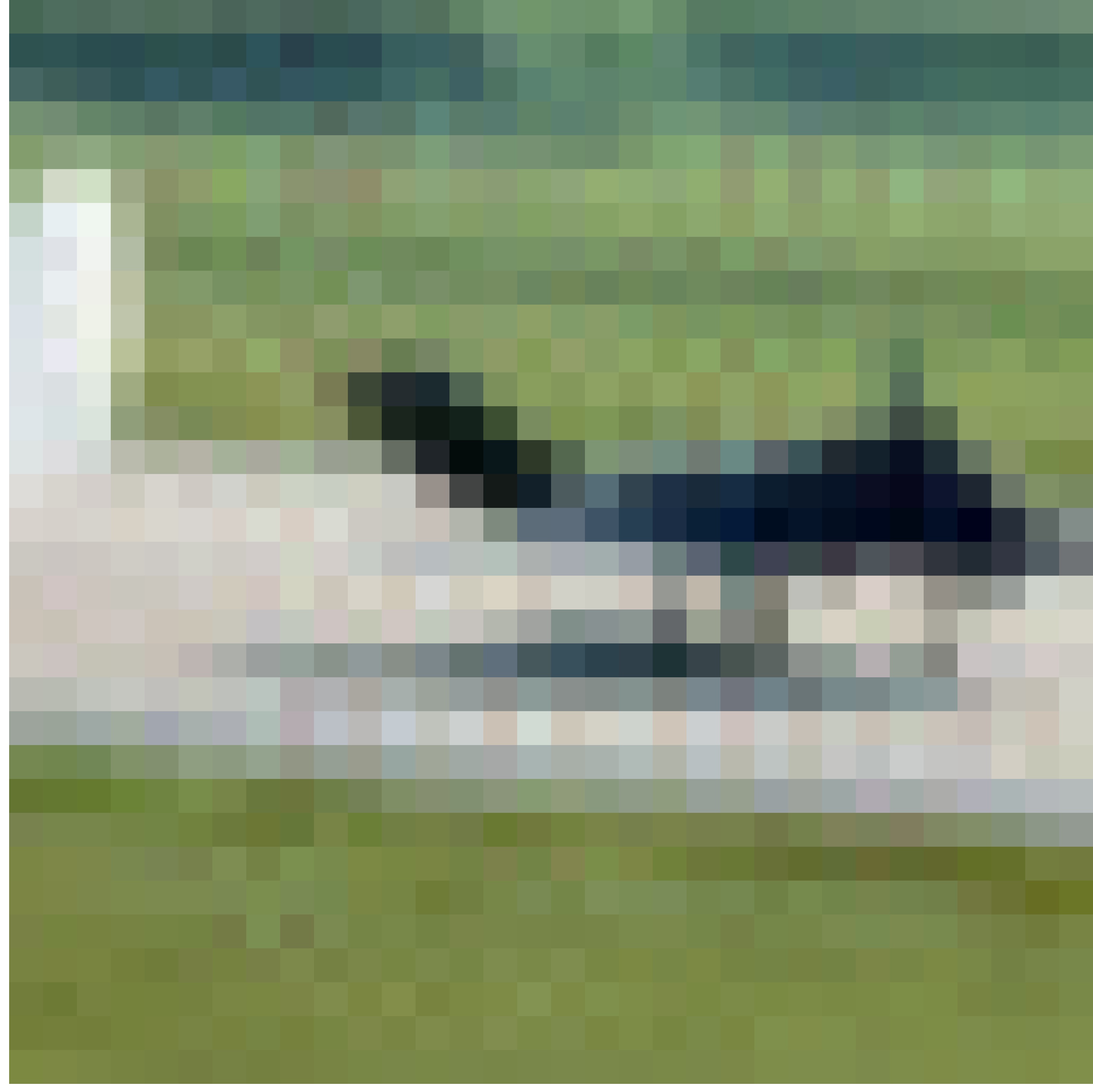}}
    \subfloat[$\epsilon_{S1}=30$]{\includegraphics[width=0.19\linewidth]{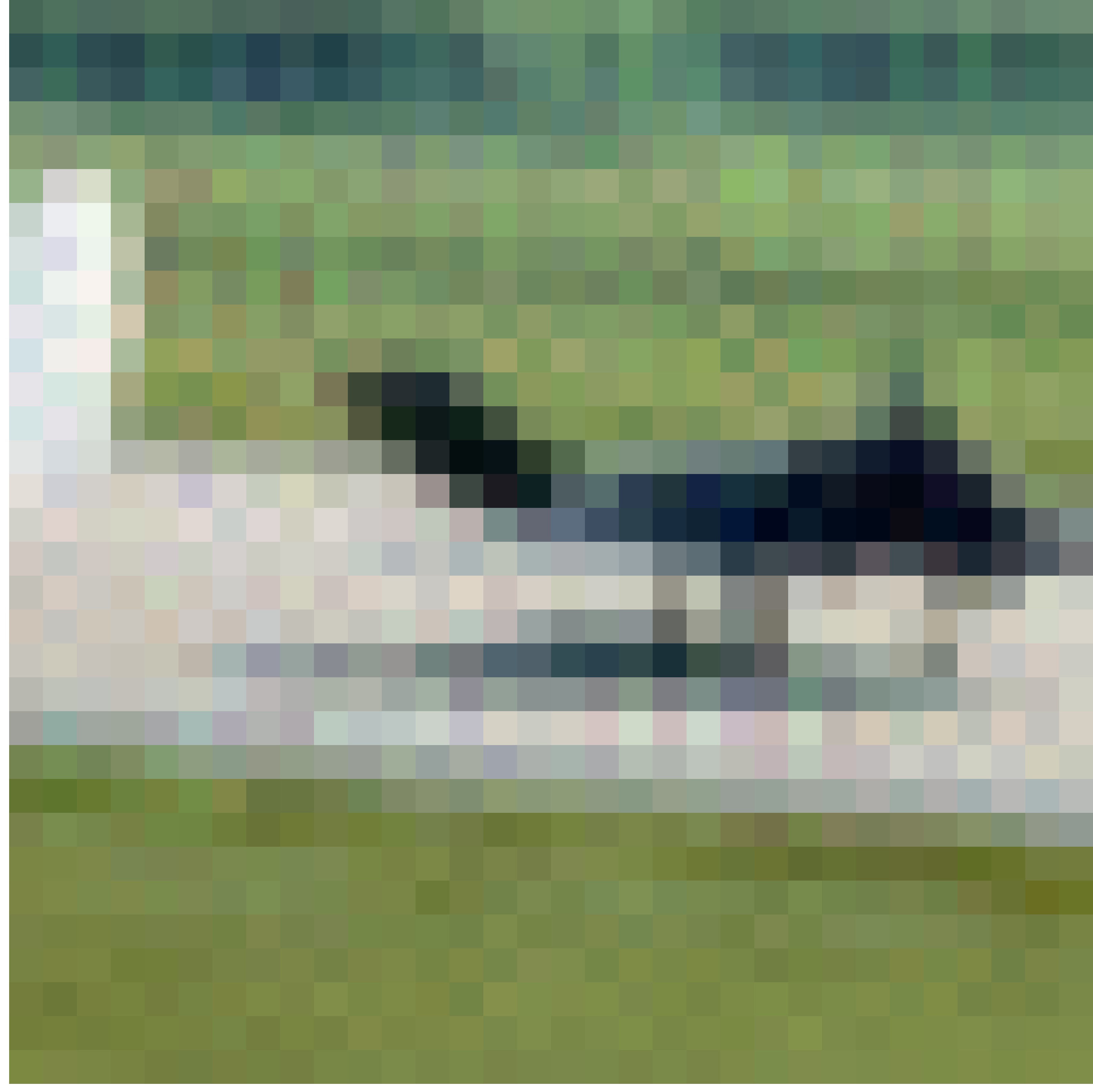}}
    
    \caption{For a test image of CIFAR-10, we computed the various adversarial examples stemming from solving \eqref{eq:opt_problem_adversarial} on the nuclear ball with Frank-Wolfe algorithm. From left to right: original image, adversarial example with a nuclear radius of $\epsilon_{S1}=5, 10, 20, 30$. Note that the adversarial examples are already miss-classified with $\epsilon_{S1}=3$; here we increase the radius purposely to observe the perturbation on the initial image.}
    \label{fig:nuclear_blur}
\end{figure}

Finally, we observe that, in practice, the adversarial perturbations add a blurring effect to the initial images, see Figure \ref{fig:nuclear_blur} for instance. In some potential security scenarios, such perturbations could then be perceived as simple alterations of the image rather than a malware deformation of it, hence not raising the human attention (see \citep{gilmer2018motivating} for real-world scenarios). 

\subsection{Group constraints}\label{ssec:group_constraints}
We now propose group-norms that depend on a partition of the pixels' coordinates into groups. For instance, such a partition can be adapted from a segmentation of the sample image. These group-norms are a combination of two norms: a local one applied on vectors formed by each group of pixels' values, and a global one applied on the vectors of the norms of all the groups. Here, we consider the nuclear norm as the local norm and the global $\ell_1$ norm to induce sparsity at the group level. Considering such norms allows to constrain the type of perturbations further.

Note here that our approach is not the only way to produce structured adversarial perturbations. However, Conditional Gradient iterates (see Section \ref{sec:FW_for_adversarial_examples}) are all structured according to the structured norm ball with no other algorithmic modification than the knowledge of the LMO. We can hence easily explore structures in adversarial examples implementing a trade-off between decrease the adversarial loss and enforcing specific structures to the perturbation.

\paragraph{Nuclear Group Norm.}
Let $\mathcal{G}$ be an ensemble of groups of pixels' coordinates of the tensor image of $(c,h,w)$. Each element $g\in\mathcal{G}$ is a set of pixel coordinates'. Then for $x\in\mathbb{R}^{c\times h \times w}$ we write, with $p\in [1,\infty[\cup\{\infty\}$,
\begin{equation}\label{eq:group_norm_schatten}
||x||_{\mathcal{G},1,p} = \sum_{g\in\mathcal{G}} ||x[g]||_{S(1)},
\end{equation}
for the $\mathcal{G}$-nuclear group-norm, see for instance \citep{tomioka2013convex}. When $\mathcal{G}$ is a partition of the pixels, $||\cdot||_{\mathcal{G},1,S(1)}$ is a norm. The nuclear group-norm allows to localize the blurring effect of the nuclear norm. Indeed, the LMO of $\mathcal{G}$-nuclear group-norm is given by
\BEQ\label{eq:LMO_group_Schatten}
  \text{LMO}_{||\cdot||_{\mathcal{G}, 1, S1}\leq \rho}(M) \triangleq \left\{
    \begin{split}
    \rho ~ U^{(g)}_1  \big(V^{(g)}_1\big)^T\\
    0~\text{ otherwise}
    \end{split}
  \right.~,
\EEQ
where $g^*=\underset{g\in\mathcal{G}}{\text{argmax }}\big|\big|M[g]\big|\big|_{S1}$ and the singular value decomposition of $M[g]$ for each group $g$ is given by $U^{(g)} S^{(g)} \big(V^{(g)}\big)^T$.
When solving \eqref{eq:opt_problem_adversarial} with such norms, each iteration of the conditional gradient will add to the adversarial perturbation a vertex of the form described by \eqref{eq:LMO_group_Schatten}, \textit{i.e.} a matrix of rank-one on the rectangle defined by the group of pixels in $g\in\mathcal{G}$.

\paragraph{Different Distortion Radius per Group.}
When perturbing an image, modification in the pixel regions with high variance are typically harder to perceive than pixels modification in low variance regions. This was leveraged on in \citep{luo2018towards} or in the $\sigma$-map of \citep[\S 2.2.]{croce2019sparse}. Weighted nuclear group norms allow to search adversarial perturbations with different distortion radius across the image. With some $w_g>0$, the weighted nuclear group norm is defined as
\BEQ
||x||_{\mathcal{G}, 1, S(1), w} = \sum_{g\in\mathcal{G}}{ w_g ||x[g]||_{S(1)}},
\EEQ
and the LMO for weighted nuclear group-norm is then obtained as
\BEQ\label{eq:LMO_group_Schatten_1}
  \text{LMO}_{||\cdot||_{\mathcal{G}, 1, S1}\leq \rho}(M) \triangleq \left\{
    \begin{split}
    \frac{\rho}{w_{g^*}} ~ U^{(g^*)}_1  \big(V^{(g^*)}_1\big)^T~\text{ on group of pixels } g^*\\
    0~\text{ otherwise}
    \end{split}
  \right.~,
\EEQ
where $g^*=\underset{g\in\mathcal{G}}{\text{argmax }}\frac{1}{w_g}\big|\big|M[g]\big|\big|_{S1}$ and the singular value decomposition of $M[g]$ for each group $g$ is given by $U^{(g)} S^{(g)} \big(V^{(g)}\big)^T$. In particular, this means that the solution corresponding to the group associated with $g$ have a nuclear radius of $\frac{\rho}{w_g}$ and the weights $w_g$ allow to control the distortion in each group of pixels.

\paragraph{Structured Attacks and Certification.}
Some defense methods aim at certifying that some neural network classifiers are constant in balls around test images \citep{wong2017provable,raghunathan2018semidefinite,raghunathan2018certified}. In particular, these methods seek at escaping the cat and mice game between empirical attacks and defenses methods. However, it is not yet clear how these scale to large datasets and neural architectures, for instance,  \citep{raghunathan2018semidefinite} only provide certifiable robustness on MNIST with $\epsilon_{\ell_{\infty}}=0.1$. 

Besides, these methods guarantee certifiable robustness in terms $\ell_p$-norms perturbations. Although norms are equivalents in the finite-dimensional space of images, their guarantee may become shallow under a different measure of the distortion set. This was pointed out with Wasserstein distortion sets in \citep[\S 5.3.]{wong2019wasserstein}.

\section{Structure Enhancing Algorithm for Adversarial Examples}\label{sec:FW_for_adversarial_examples}
We apply the Frank-Wolfe algorithms \citep{frank1956algorithm}, a.k.a. conditional gradient algorithms \citep{levitin1966constrained}, for problem \eqref{eq:opt_problem_adversarial}. These algorithms have known a recent revival in constrained optimization problems for machine learning. This is notably due to their low cost computational cost per iterations \citep{jaggi2013revisiting} as well as the many related theoretical and practical open questions, like linear convergence on polytopes \citep{guelat1986some,garber2013linearly,garber2013playing,
lacoste2013affine,FW-converge2015}, convergence on strongly convex set \citep{levitin1966constrained,demyanov1970,dunn1979rates,garber2015faster} or uniformly convex sets \citep{kerdreux2020frank}.

On specific constraint structures, such as the one developed in Section \ref{sec:structured_distortion_sets}, conditional gradient algorithms naturally trade off the convergence accuracy with the structure of the early iterates. Note also that for the case of large-scale nuclear norm regularization in convex optimization, the Frank-Wolfe algorithm has been extensively studied \citep{jaggi2010simple,lee2010practical,shalev2011large,harchaoui2012large,dudik2012lifted,allen2017linear,garber2018fast}. Many variations \citep{freund2017extended,cheung2017projection} exist which leverage the facial properties (see \citep[Theorem 3]{freund2017extended} or originally from \citep{so1990facial}) of the nuclear ball (which is not a polytope nor a strongly convex set).

\begin{algorithm}[h!]
   \caption{Vanilla Frank-Wolfe}
   \label{alg:vanilla_FW}
\begin{algorithmic}
   \STATE {\bfseries Input:} Original image $\xx_0$
    \FOR{$t=0,\cdots, T$}
        \STATE $s_{t} = \text{LMO}_{\mathcal{C}}\big(-\nabla f(\xx_t)\big)$.
        \STATE $\gamma_t=\text{LineSearch}(x_{t}, s_t-x_t)$
        \STATE $x_{t+1} = (1 - \gamma_t)x_{t} + \gamma_t s_{t}$
   \ENDFOR
\end{algorithmic}
\end{algorithm}

For all the distortion set we consider, the LMO is explicit. While we do not focus on computational efficiency, we note that the computation of the LMO has a low computational requirement as opposed to projection based approaches. Indeed it require only the first singular vectors as opposed to proximal steps which require the full SVD. 
Provided access (when applicable) to an upper-bound $L$ of the adversarial loss Lipschitz constant in \eqref{eq:opt_problem_adversarial}, we use the short step size rule $\gamma_t=\text{clip}_{[0,1]}(\langle -\nabla f(x_{t}), s_t-x_t\rangle/{L ||s_t - x_t||^2})$. \cite{chen2018frank} consider using zero-order Frank-Wolfe algorithm for solving adversarial problems like \eqref{eq:opt_problem_adversarial} but in the black-box setting.

When the objective functions are non-convex, \textit{e.g.} with adversarial losses, injecting noise may be beneficial. For instance, this can be done either via random starts or via randomized block-coordinate methods. With some restriction \cite{kerdreux2018frank} propose a version of Frank-Wolfe that solves linear minimization oracles on random subsets of the constraint sets. For the nuclear group norm, the sampling scheme could be done at the group-level. For instance, we consider the nuclear group norm with one group per channel, \textit{i.e.} $||x||_{color, S1}= \sum_{c=1}^{3}||x_c||_{S1}$ where $x_c$ is one of the image channels. We experiment this approach in Section \ref{sec:numerical_experiments}.

Here, we did not consider the constraints that the images iterates should belong to the $[0,1]^d$ box constraints. Instead, we clamp the last iterate to belong to the box constraints. This does not guarantee the convergence to a saddle point but removes the need to compute the Linear Minimization Oracle over the intersection of two sets, which is non-trivial. We are ultimately interested in the effective success rate of the attack that we explore in the subsequent section.

\section{Numerical Experiments}\label{sec:numerical_experiments}
This section aims at evaluating the success rate in different scenarios of adversarial examples stemming from the application of Frank-Wolfe algorithms to the adversarial problem \eqref{eq:opt_problem_adversarial} with (group) nuclear balls as distortion sets, which we refer as FWnucl. 

\paragraph{Experiments Goal.} We tested FWnucl white-box attack against two baselines of defenses for untargeted attacks. The first is \cite{madry2017towards}, the state-of-the-art defense against white-box attacks. It uses the training images augmented with adversarial perturbations to train the network.

The second one \cite{yang2019me} leverages matrix estimation techniques as a pre-processing step; each image is altered by randomly masking various proportions of the image pixels' and then reconstructed using matrix estimation. For a given training image, this produces a group of images that are used during training, see \citep[\S 2.3.]{yang2019me} for more details. This provides a non-differentiable defense technique, \textit{i.e.} a method that cannot be straightforwardly optimized via back-propagation algorithms, and was reported to be robust against methods in \citep{athalye2018obfuscated} by circumventing the obfuscated gradients defenses. Qualitatively it leverages a structural difference between the low-rank structure of natural images and the adversarial perturbations that are not specifically designed to share the same structures.
A key motivation of our work then is to propose adversarial examples with specific structures, serving at least as a sanity check for defense approaches in the spirit of \citep{yang2019me}.

We report the attack success rates of FWnucl along with those of classical attack methods like Fast Gradient Sign Method (FGSM) \cite{goodfellow2014explaining}, and Projected Gradient Descent (PGD) \cite{madry2017towards}, both of which being different methods to solve the same adversarial problem \eqref{eq:opt_problem_adversarial} where the distortion set is the $\ell_{\infty}$ ball. For each technique, we report accuracy as the percentage of adversarial examples that are classified correctly. These numerical experiments demonstrate that the attack success rates for FWnucl are comparable to the classical ones in an imperceptibility regime while also retaining specific structures in the perturbation.

\paragraph{Experiment settings.} We assess accuracy of networks on MNIST and CIFAR-10 testsets. For ImageNet we randomly selected $4000$ from  the ImageNet validation set that is correctly classified.
As classically done in previous works, for MNIST, we use the LeNet model with two convolutional layers similar to \cite{madry2017towards} and SmallCNN with four convolutional layers followed by three fully connected layers as in \cite{carlini2017towards}. For CIFAR-10 dataset we use ResNet-18 and its wide version WideResNet and ResNet-50. 

\paragraph{Nuclear Attacks Structures.}
In MNIST dataset there is no texture besides the digits' active pixels representing the figures. In particular, attacks that tend to perturb all pixels are not good candidates as they require low distortion parameters to be imperceptible. This is, for instance, the case of the Frank-Wolfe attack with nuclear ball distortion sets. Indeed nuclear adversarial examples perturb nearly all the pixels, with values ranging from 1 to 5 (with respect to 255). This is because at each iteration Frank-Wolfe algorithms add a rank-one matrix to the perturbation. These rank one matrices stem from the right and left singular vector of the initial matrix and generally involve many of the image pixels.

\begin{table}
\centering
\caption{MNIST and CIFAR-10 extensive white-box attack results. FWnucl 20$\,^{*}$: FWnucl with $\epsilon_{S1} = 1$. FWnucl 20$\,^{+}$: FWnucl with $\epsilon_{S1} = 3$. On MNIST (resp. CIFAR-10) PGD and FGSM have a total perturbation scale of 76.5/255(0.3) (resp. 8/255 (0.031)), and step size 2.55/255(0.01) (resp. 2/255(0.01)). PGD runs for 20 iterations. We reproduce the ME-Net and Madry defense with same training hyper-parameters.}
\label{tab:attack_acc}
\begin{tabular}{@{\hspace{1cm}}ll ccccc}
\toprule
\multirow{2}{*}{\textbf{Network}} & \multirow{2}{*}{\textbf{Training}}  & \multirow{2}{*}{Clean} &
\multicolumn{4}{c}{\textbf{Accuracy under attack}}
\\\cmidrule{4-7}
 & {\textbf{Model}} &  & FWnucl 20$\,^{*}$ & FWnucl 20$\,^{+}$ & PGD 20 &  FGSM  \\
\midrule
\multicolumn{6}{l}{\textbf{MNIST}}\\\cmidrule{1-1}
\multirow{2}{*}{LeNET} & {Madry} & 98.38& {\bf 95.26} & {\bf 92.76} & 95.79 & 96.59 \\
& {ME-Net} & 99.24 & 97.63 & 75.41 & 74.88 & {\bf  46.18} \\
\addlinespace
 \midrule
 \multirow{2}{*}{SmallCNN} & {Madry} & 99.12 & 98.19 & 96.66 & {\bf 95.77} & 97.95 \\
& {ME-Net} & 99.42 & 89.56 & 78.65 & 76.84 & {\bf 54.09} \\
\addlinespace
 \midrule
 \midrule
\multicolumn{6}{l}{\textbf{CIFAR-10}}\\\cmidrule{1-1}
\multirow{2}{*}{ResNet-18} & {Madry} & 81.25 & {\bf 44.28} & {\bf 3.06} & 49.95 & 55.91 \\
& {ME-Net} & 92.09 & 29.66 & {\bf 4.01} & 4.99 & 44.80 \\
\addlinespace
\midrule
\multirow{2}{*}{WideResNet} & {Madry} & 85.1 & {\bf 43.16} & {\bf 2.82} & 52.49 & 59.06 \\
& {ME-Net} & 92.09 & 40.09 & 16.04 & {\bf 12.73} & 59.33 \\
\midrule
\multirow{2}{*}{ResNet-50} & {Madry} & 87.03 & {\bf 40.97} & {\bf 2.64} & 53.01 & 61.44  \\
& {ME-Net} &  92.09 & 47.66 & 17.81 & {\bf 9.14} & 58.51\\
\addlinespace
\bottomrule
\end{tabular}
\end{table}

In Table \ref{tab:norms_cifar} we report the mean $\ell_2$, and nuclear norms of the adversarial noise over all attacks in Table \ref{tab:attack_acc} for the CIFAR-10 dataset (see the Appendix for MNIST dataset). Our method with $\epsilon_{S1} = 1$ generates perturbations with almost $7x$ and $10x$ lower $\ell_2$ norm for the MNIST dataset. Interestingly, the adversarial examples for FWnucl show significantly lower nuclear norm.

\begin{table}
\centering
\caption{Comparison of the white-box attacks for  CIFAR-10 on ResNet-18 adversarially trained. PGD, on the $\ell_{\infty}$ ball, and FGSM have a total perturbation scale of 8/255 (0.031), and step size 2/255(0.01).
FWnucl 20$\,^{*}$: FWnucl with $\epsilon_{S1} = 1$. FWnucl 20$\,^{+}$: FWnucl with $\epsilon_{S1} = 3$.
}
\label{tab:norms_cifar}
\begin{tabular}{@{\hspace{1cm}}lc ccc ccc}
\toprule
&\multicolumn{2}{c}{\textbf{ResNet-18}}&\multicolumn{2}{c}{\textbf{WideResNet}}&\multicolumn{2}{c}{\textbf{ResNet-50}}
\\\cmidrule{2-7}
{{Attack}} & Mean $\ell_2$ & Mean $\norm{\cdot}_{{S1}}$ & Mean $\ell_2$ & Mean $\norm{\cdot}_{{S1}}$ & Mean $\ell_2$ & Mean $\norm{\cdot}_{{S1}}$\\
\midrule
{FWnucl 20$\,^{*}$} & {\bf 1.38} & {\bf 0.91} & {\bf 1.36} & {\bf 0.90} & {\bf 1.31} & {\bf 0.91}  \\ 
{FWnucl 20$\,^{+}$} & 3.37 & 2.72 &3.24 & 2.62 &3.00 & 2.65 \\
{PGD 20}& 1.68 & 3.88 & 1.68 & 3.97 & 1.66 &3.89\\
{FGSM}& 1.73 & 4.04 &  1.73 & 4.04 & 1.73 & 4.10 \\
\bottomrule
\end{tabular}
\end{table}

\begin{figure}
\subfloat{
\centering
\includegraphics[width=0.44\linewidth]{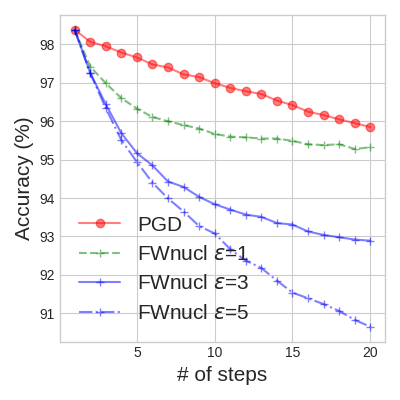}}%
\subfloat[]{
\centering
\includegraphics[width=0.44\linewidth]{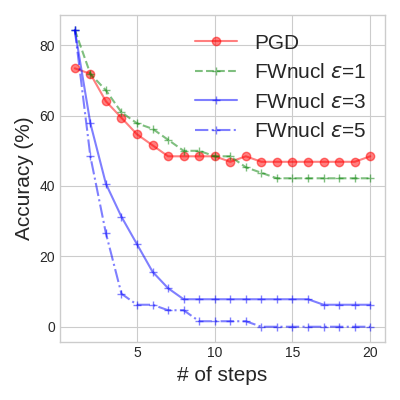}}%
\setlength{\abovecaptionskip}{2pt}
\caption{Accuracy of robust models, MNIST (left) and CIFAR-10 (right), versus the number of steps in PGD and FWnucl attacks when varying the nuclear ball radius on the latter.}
\label{fig:attack_FW_increase_eps}
\end{figure}

In Figure \ref{fig:attack_FW_increase_eps} we report the accuracy of adversarially trained networks over MNIST and CIFAR-10 datasets when attacked with FWnucl. It shows the network's accuracy at each iteration of a FWnucl attack with various nuclear norm radius $\epsilon_{S1} = 1, 3, 5$.

\begin{figure}
\subfloat{
\centering
\includegraphics[width=0.44\linewidth]{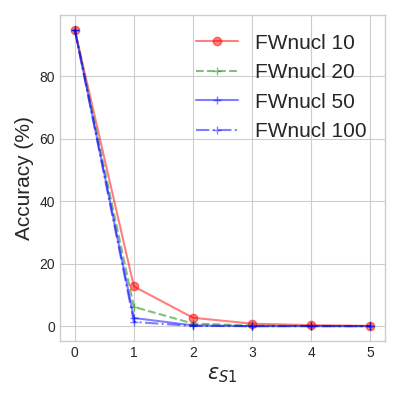}}%
\subfloat[]{
\centering
\includegraphics[width=0.44\linewidth]{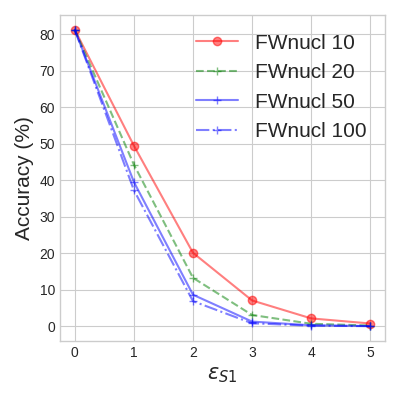}}%
\setlength{\abovecaptionskip}{2pt}
\caption{Accuracy of standard model (left) and robust model of Madry (right) on ResNet-18 for CIFAR-10, versus the nuclear ball radius when varying the number of steps.} 
\label{fig:attack_FW_radius_increase_steps}
\end{figure}

Figure \ref{fig:attack_FW_radius_increase_steps} summarizes the results for FWnucl with varying for standard and robust model on CIFAR-10. The FWnucl algorithm noticeably drops the accuracy rate by increasing the radius $\epsilon_{S1}$. The performance of different FWnucl methods is slightly different, as more FWnucl steps may gain better performances.

\begin{figure}
\subfloat[]{
\centering
\includegraphics[width=0.90\linewidth]{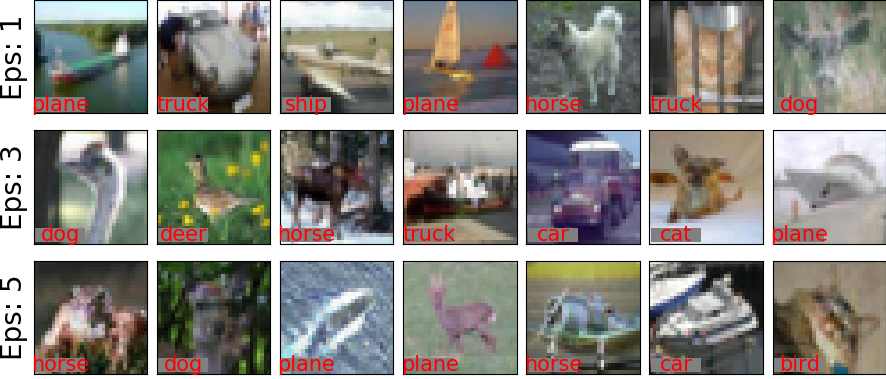}}%
\setlength{\abovecaptionskip}{2pt}
\caption{FWnucl adversarial examples for the CIFAR-10 dataset for different radii.  The fooling label is shown on the image.}
\label{fig:mnist-cifar-images}
\end{figure}

\begin{figure}
\centering
\subfloat[Bathtub]{{\includegraphics[scale=0.22]{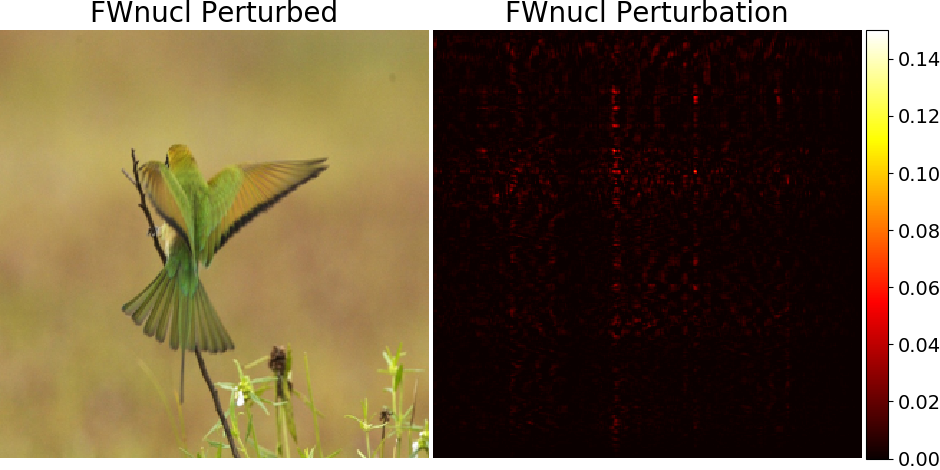}}}
\subfloat[Basset]{{\includegraphics[scale=0.22]{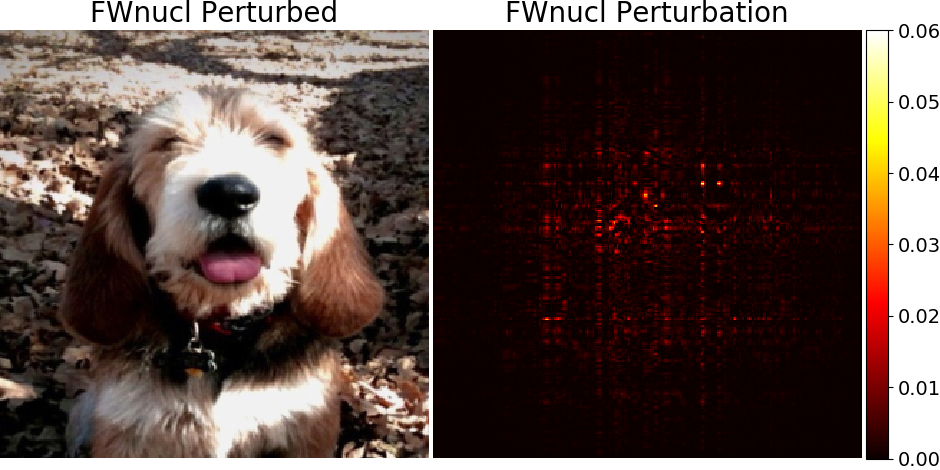}}}
\subfloat[Bouvier des Flandres]{{\includegraphics[scale=0.22]{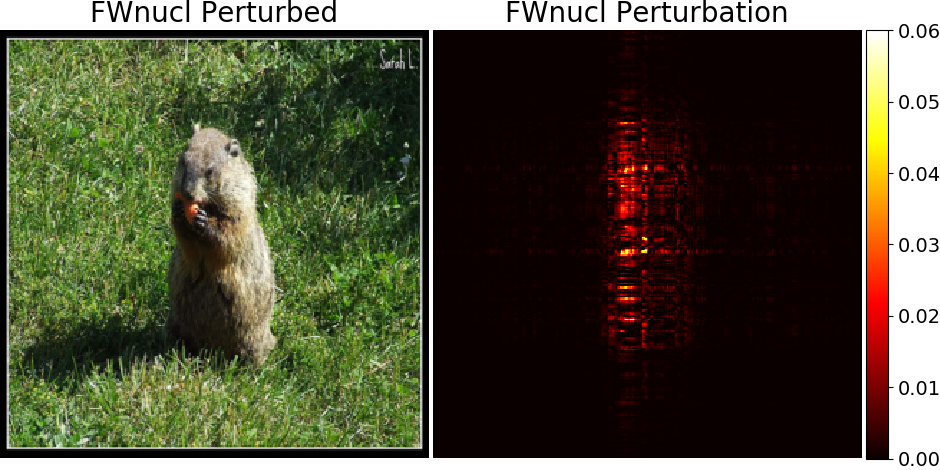}}}

\subfloat[Golf Ball]{{\includegraphics[scale=0.22]{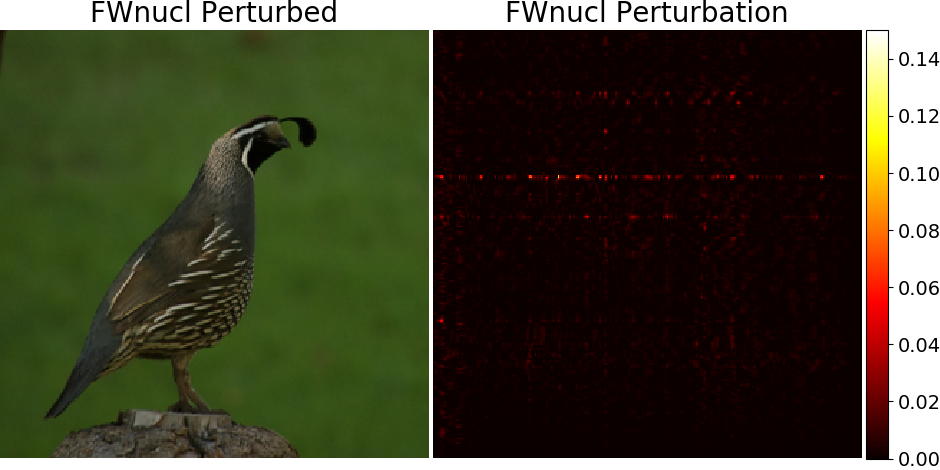}}}
\subfloat[Hog]{{\includegraphics[scale=0.22]{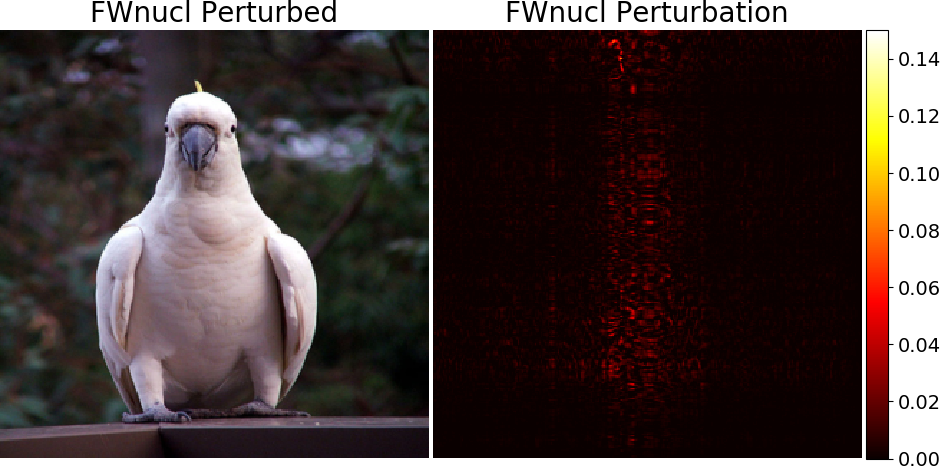}}}
\subfloat[Custard Apple]{{\includegraphics[scale=0.22]{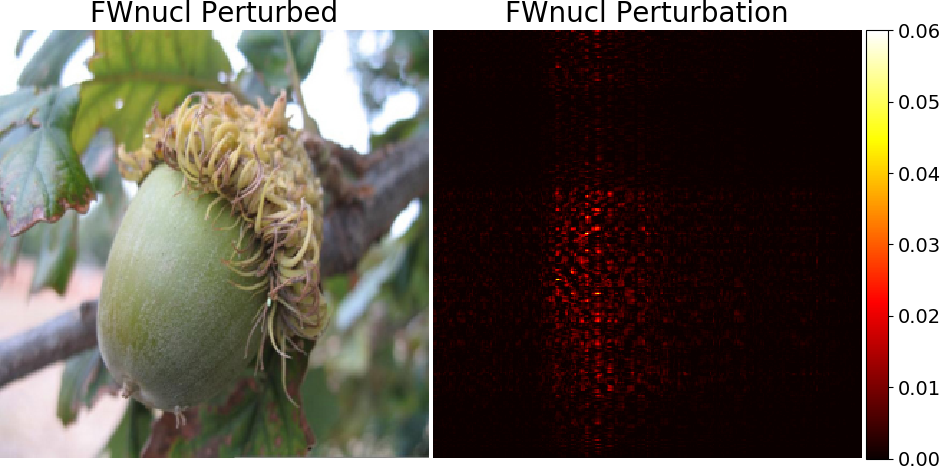}}}

\caption{The images display some structural pattern of FWnucl perturbations for the ImageNet dataset on DenseNet121 architecture for various level of distortion, standardly trained. Observe that the adversarial perturbed pixels are accumulated on the areas containing semantic information about the image. FWnucl is conducted with $\epsilon_{S1}=5$ and $20$ iterations.}\label{fig:distortion_imagenet}
\end{figure}

\paragraph{Imperceptibility nuclear threshold.}
We illustrate in Figure \ref{fig:mnist-cifar-images} some adversarial examples generated by FWnucl, for three different values of epsilon. The imperceptibility threshold depends on the dataset. On CIFAR-10, we qualitatively observed that with $\epsilon_{S1} = 1 $, all adversarial examples are perceptually identical to the original images. Also as the dataset becomes more complex, the tolerance of imperceptibility to nuclear ball radius values $\epsilon_{S1}$ increases; on ImageNet the imperceptibility threshold is qualitatively for $\epsilon_{S1}=10$.
In Figure \ref{fig:distortion_imagenet}, we observe that the perturbations are particularly congregated around important regions (i.e., body, head), although there is not a universal configuration to detect specific features that are the  most  important  for  the  network. 
While the noise generated by PGD attack exhibits abrupt changes in pixel intensities (see Figure \ref{fig:example_diff_distortion}), the perturbation from FW has a continuous variations in pixel values. In addition, the number of non-zero pixel values for PGD and FGSM on ImageNet is  almost $11 x$ and $14 x$ larger, respectively than the number of non-zero pixel intensities for FWnucl with $\epsilon_{S1}=1$.

It is important to characterize the type of deformation that arise with radii above the imperceptibility threshold as the imperceptibility regimes are not the only security scenario. In particular accuracy of robust networks quickly drop to zero in these regimes, see Figure \ref{fig:attack_FW_increase_eps}-\ref{fig:attack_FW_radius_increase_steps}, facilitating the attacks. In the nuclear ball case, as the radius $\epsilon_{S1}$ of the nuclear ball increases, the perturbation becomes perceptible with a blurring effect. Structure in the adversarial examples can be leveraged to create specific perceptible deformation effects that look natural to humans.

\begin{figure}[H]
    \centering
    \subfloat[ResNet-18]{{\includegraphics[scale=0.22]{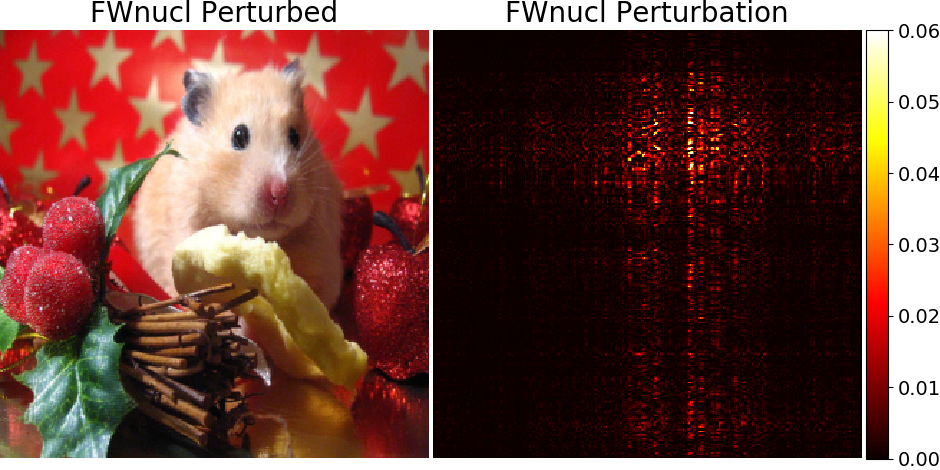}}}
    \subfloat[DenseNet121]{{\includegraphics[scale=0.22]{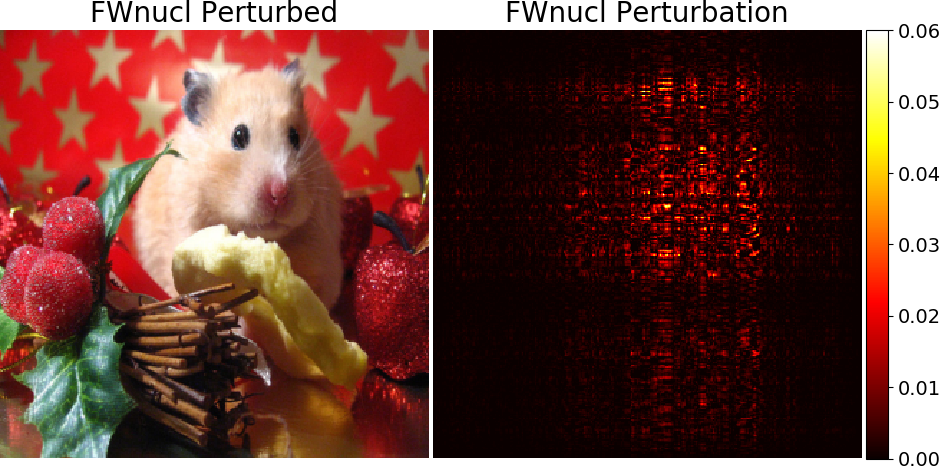}}}
    \subfloat[GoogLeNet]{{\includegraphics[scale=0.22]{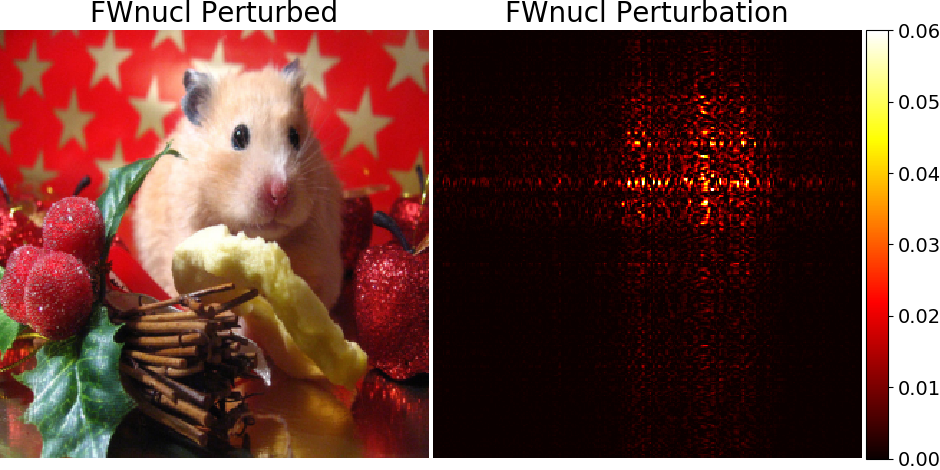}}}
    \caption{General layout of the FWnucl perturbations for ImageNet across three different architectures.}
   \label{fig:trans_imagenet_ham}
\end{figure}

\paragraph{Transferability.} In Table \ref{tab:trans_imagenet} we investigate the transferability of FWnucl adversarial examples over different architectures for ImageNet.
\begin{table}
\centering
\caption{Fooling rates of FWnucl adversarial perturbations between several models for 4000 samples from ImageNet. The row indicates the source model and the column indicates the target model.}\label{tab:trans_imagenet}
\begin{tabular}{|c|c|c|c|}
\cline{2-4}
\multicolumn{1}{c|}{} & ResNet-18 & DenseNet121 & GoogLeNet \\ \hline
ResNet-18 & 100  & 18.15   & 12.91  \\ \hline
DenseNet121 & 16.56   & 99.30   & 11.74  \\ \hline
GoogLeNet & 15.03   & 12.37   & 99.40  \\ \hline
\end{tabular}
\end{table}
This table shows that there should be some similar structural pattern between independent architectures that FWnucl employs, but the adversaries are mainly network dependent.
In Figure \ref{fig:trans_imagenet_ham}, we illustrate how the adversarial nuclear structure vary from one network to another for the same image; in particular the perturbation continuously concentrate around the important regions of the image with however varying layout and pattern of perturbation for each network.

\section{Conclusion}
We consider adversarial attacks beyond  $\ell_p$ distortion set. We propose a simple optimization approach producing structured adversarial examples with non-$\ell_p$ distortion sets. First, it challenges the theoretical certification approaches that so far are given in $\ell_p$ norm terms. Also, it allows an attacker to design perceptible adversarial examples with specific characteristics, like localized blurriness. Finally, in the imperceptible regime, some defensive pre-processing techniques may rely on a lack of certain patterns in the adversarial perturbations to destroy it. Evaluating robustness against various structured adversarial examples then seems to be a reasonable defense sanity check.

\newpage

\section*{Acknowledgement}
T.K. would like to thanks Geoffrey Negiar for interesting discussions on the topic.

%
%
\bibliographystyle{plainnat}
\bibliography{mainbib}
\newpage
\appendix


\title{Supplementary Materials}
\maketitle

\section{More on structured norms}\label{app:more_on_norms}

In \citep[\S 2.3.]{amini2017nonsmooth} introduce $\Sigma_{\mathcal{G}}$ a $1$-group-$p$-Schatten norms where the group can overlaps. In this case also it is possible to access the LMO leveraging on the knowledge of the dual norm $\Sigma_{\mathcal{G}}$
\[
\Sigma_{\mathcal{G}}^{\circ}(M) = \underset{g\in\mathcal{G}}{\text{max }} \big|\big| M[g]
\big|\big|_{S(q)},
\]
where $S(q)$ with $1/p + 1/q=1$.

From \citep[(2.5),(2.6)]{pierucci2017nonsmooth} shows that the group-nuclear norms $||\cdot||_{\mathcal{G}, 1 ,S(1)}$ are convex surrogates of some group-rank function and as such enforce solution that are low-rank on some groups.

We stated the LMO $q$-Schatten for $q=1$, which corresponds to the trace-norm or nuclear norm, and $q\in]1, +\infty[$ separately. The LMO for the Schatten norm with $q=+\infty$ is given by
\BEQ\label{eq:sol_schatten_inf_LMO}
\text{LMO}_{||\cdot||_{S({\infty})}\leq \rho} \triangleq \rho U V^T,
\EEQ
where $M=U S V^T$ with $U$ and $V$ the matrix of left and right normalized singular vectors.\\

So far we only considered sparsity (at the group level) inducing group-norm, but more general norms can be simply written with $r\in]1,+\infty[$ and $p\in[1,+\infty[$
\begin{equation}\label{eq:group_norm_schatten_general}
||x||_{\mathcal{G},r,S(p)} = \Big(\sum_{g\in\mathcal{G}} ||x[g]||^r_{S(p)}\Big)^{1/r}~,
\end{equation}
however, sparsity of the global adversarial perturbation is something which is not interesting to loose as it concentrates the perturbation only on some areas.


\section{More on numerical results}\label{app:more_numerical_results}

\begin{figure}
    \centering
    \subfloat[$\epsilon_{S1}=3$]{
    \centering
    \includegraphics[width=0.35\linewidth]{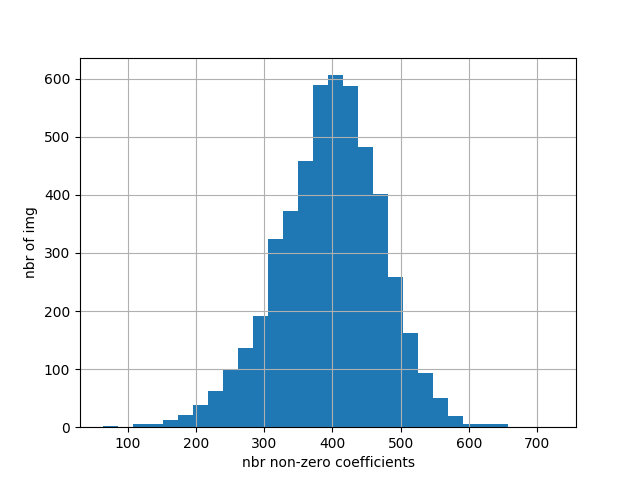}}
    \subfloat[$\epsilon_{S1}=5$]{
    \centering
    \includegraphics[width=0.35\linewidth]{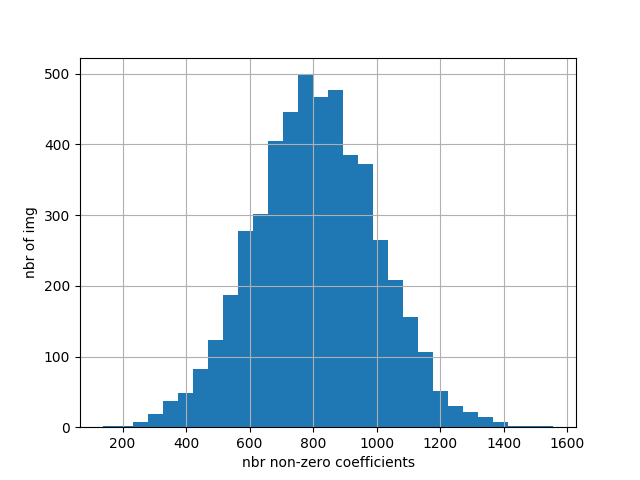}}
    \caption{Showing the distribution of number of modified pixels between a CIFAR10 test image with integer pixel values in $[0,255]$ and a perturbed image in the same format. We use $5000$ images in CIFAR10 test set. Each image is perturbed via adding $\epsilon_{S_1} u v^T$ to the normalized initial test image and then by unnormalizing and clamping it to integers values between $[0, 255]$. $u$ and $v$ are the right and left singular vectors associated to the largest singular value.
    In Frank-Wolfe nuclear the Frank-Wolfe vertex that are added to the perturbation are of the form of $\epsilon_{S_1} u v^T$ where $(u,v)$ are singular-vectors associated to the largest singular value of the perturbed image along the iterations.}
    \label{fig:rank_one_from_img_and_pixels}
\end{figure}


\begin{table}[H]
\centering
\caption{MNIST and CIFAR-10 extensive white-box attack results against standardly trained models. FWnucl 20$\,^{*}$: FWnucl with $\epsilon = 1$. FWnucl 20$\,^{+}$: FWnucl with $\epsilon = 3$.}
\label{tab:attack_acc_std}
\begin{tabular}{@{\hspace{1cm}}ll ccccc}
\toprule
\multirow{2}{*}{\textbf{Network}} & \multirow{2}{*}{Clean} &
\multicolumn{4}{c}{\textbf{Accuracy under attack}}
\\\cmidrule{3-7}
 &  & FWnucl 20$\,^{*}$ & FWnucl 20$\,^{+}$ & PGDnucl 20& PGD 20 &  FGSM \\
\midrule
\multicolumn{6}{l}{\textbf{MNIST}}\\\cmidrule{1-1}
\multirow{1}{*}{LeNET} & 99.32 & 93.82& 36.24 & 98.01 & 0.25 & 38.62 \\
\addlinespace
 \midrule
 \multirow{1}{*}{SmallCNN} & 99.47 & 94.21 & 76.9 & 97.32  & 20.02 & 90.63\\
\addlinespace
 \midrule
 \midrule
\multicolumn{6}{l}{\textbf{CIFAR-10}}\\\cmidrule{1-1}
\multirow{1}{*}{ResNet-18} & 93.94 & 2.77 & 0.00 & 24.22 & 0.00 & 18.60\\
\addlinespace
\midrule
\multirow{1}{*}{WideResNet} & 95.28 & 3.65 & 0.00 & 25.18 & 0.00 & 26.86\\
\midrule
\multirow{1}{*}{ResNet-50} & 93.22 & 2.99 & 0.00 & 19.35 & 0.00 & 22.26 \\
\addlinespace
\bottomrule
\end{tabular}
\end{table}

\begin{table}\hfill
\centering
\caption{MNIST and CIFAR-10 extensive white-box attack results with $\epsilon_{S1} = 5$.}
\label{tab:attack_eps5}
\begin{tabular}{@{\hspace{1cm}}ll ccccc}
\toprule
\multirow{2}{*}{\textbf{Network}} & \multirow{2}{*}{\textbf{Training}}  & & \multicolumn{4}{c}{\textbf{Accuracy under attack}}
\\\cmidrule{3-7}
 & {\textbf{Model}} &  & FWnucl 10 & FWnucl 20 & FWnucl 50 & FWnucl 100 \\
\cmidrule{1-7}
\multicolumn{6}{l}{\textbf{MNIST}}\\\cmidrule{1-1}
\multirow{2}{*}{LeNET} & {Madry} & & 92.67 & 90.39 & 87.84 & 86.61 \\
& {ME-NET} & & 50.60 & 39.02 & 30.12 & 26.65 \\
\addlinespace
\cmidrule{1-7}
 \multirow{2}{*}{SmallCNN} & {Madry} & & 96.34 & 94.5 & 92.17 & 91.02 \\
& {ME-NET} & & 63.75 & 60.98 & 55.98 & 54.65 \\
\addlinespace
 \cmidrule{1-7}
 \cmidrule{1-7}
\multicolumn{6}{l}{\textbf{CIFAR-10}}\\\cmidrule{1-1}
\multirow{2}{*}{ResNet-18} & {Madry} & & 0.74 & 0.19 & 0.04 & 0.01 \\
& {ME-NET} & & 3.21 & 0.80 & 0.29 & 0.10 \\
\addlinespace
\cmidrule{1-7}
\multirow{2}{*}{WideResNet} & {Madry} & & 0.81 & 0.19 & 0.03 & 0.02 \\
& {ME-NET} & & 14.19 & 8.32 & 4.42 & 3.29 \\
\cmidrule{1-7}
\multirow{2}{*}{ResNet-50} & {Madry} & & 1.09 & 0.25 & 0.06 & 0.03 \\
& {ME-NET} & & 16.40 & 9.17 & 4.28 & 2.89 \\
\addlinespace
\cmidrule{1-7}
\end{tabular}
\end{table}



\begin{table}
\centering
\caption{Comparison of the white-box attacks for  MNIST on SmallCNN adversarially trained. PGD, on the $\ell_{\infty}$ ball, and FGSM have a total perturbation scale of 76.5/255(0.3), and step size 2.55/255(0.01). PGD runs for 20 iterations.}
\label{tab:norms_mnist}
\begin{tabular}{@{\hspace{1cm}}lc ccc c}
\toprule
&\multicolumn{2}{c}{\textbf{LeNet}}&\multicolumn{2}{c}{\textbf{SmallCNN}}
\\\cmidrule{2-5}
{{Attack}} & Mean $\ell_2$ & Mean $\norm{\cdot}_{{S2}}$ & Mean $\ell_2$ & Mean $\norm{\cdot}_{{S2}}$ \\
\midrule
{FWnucl 20 ($\epsilon_{S1} = 1$)} & {\bf 0.45} & {\bf 0.75} & {\bf 0.45} & {\bf 0.80}  \\ 
{FWnucl 20 ($\epsilon_{S1} = 3$)} & 0.93 & 1.91 & 1.00 & 2.14  \\
{PGD 20}& 3.87 & 16.12 & 4.67 & 18.98    \\
{FGSM}& 6.15 & 24.92 & 3.57 & 12.17  \\
\bottomrule
\end{tabular}
\end{table}

\begin{table}[H]
\centering
\caption{Fooling rates of FWnucl adversarial perturbations between several models for CIFAR-10 test set. The row indicates the source models on which adversarial examples are grafted and the column indicates the target model on which these attacks are evaluated. These models are not adversarially trained. FWnucl run for $20$ iterations with $\epsilon_{S1}=5$.}\label{tab:trans_cifar}
\begin{tabular}{|c|c|c|c|}
\cline{2-4}
\multicolumn{1}{c|}{} & ResNet-18 & DenseNet121 & GoogLeNet \\ \hline
ResNet-18 & 97.48   & 86.00   & 84.64  \\
 \hline
DenseNet121 & 91.31   & 98.36   & 90.45  \\ \hline
GoogLeNet & 83.55   & 86.03   & 98.37 \\
 \hline
\end{tabular}
\end{table}

\begin{table}\hfill
\centering
\caption{Comparisons of the attacks for projected gradiet descent with nuclear norm (PGDnucl). MNIST and CIFAR-10 extensive white-box attack results for PGDnucl. PGDnucl 20$\,^{*}$: PGDnucl 20 with $\epsilon_{S1} = 1$; PGDnucl 20$\,^{+}$: PGDnucl 20 with $\epsilon_{S1} = 3$; PGDnucl 20$\,^{\#}$: PGDnucl 20 with $\epsilon_{S1} = 5$}
\label{tab:attack_pgd_nucl}
\begin{tabular}{@{\hspace{1cm}}ll ccccc}
\toprule
\multirow{2}{*}{\textbf{Network}} & \multirow{2}{*}{\textbf{Training}}  & & \multicolumn{3}{c}{\textbf{Accuracy under attack}}
\\\cmidrule{3-6}
 & {\textbf{Model}} &  & PGDnucl 20$\,^{*}$ & PGDnucl 20$\,^{+}$ & PGDnucl 20$\,^{\#}$ \\
\cmidrule{1-6}
\multicolumn{6}{l}{\textbf{MNIST}}\\\cmidrule{1-1}
\multirow{2}{*}{LeNET} & {Madry} & & 95.36 & 87.48 & 69.50  \\
& {ME-NET} & & 98.21 & 97.83 & 94.78  \\
\addlinespace
\cmidrule{1-6}
 \multirow{2}{*}{SmallCNN} & {Madry} & & 98.15 & 93.20 & 80.51 \\
& {ME-NET} & & 88.35 & 85.33 & 76.23  \\
\addlinespace
 \cmidrule{1-6}
 \cmidrule{1-6}
\multicolumn{6}{l}{\textbf{CIFAR-10}}\\\cmidrule{1-1}
\multirow{2}{*}{ResNet-18} & {Madry} & & 78.38 & 74.46 & 58.57  \\
& {ME-NET} & & 87.21 & 76.28 & 58.58  \\
\addlinespace
\cmidrule{1-6}
\multirow{2}{*}{WideResNet} & {Madry} & & 81.49 & 76.38 & -- \\
& {ME-NET} & & 89.08 & 74.74 & 54.37 \\
\cmidrule{1-6}
\multirow{2}{*}{ResNet-50} & {Madry} & & 82.87 & 77.19 & 60.68 \\
& {ME-NET} & & 87.26 & 78.29 & 62.90 \\
\addlinespace
\cmidrule{1-6}
\end{tabular}
\end{table}

\begin{table}[H]
\centering
\caption{MNIST and CIFAR-10 extensive white-box attack results with random initialization. FWnucl 20$\,^{*}$: FWnucl with $\epsilon = 1$. FWnucl 20$\,^{+}$: FWnucl with $\epsilon = 3$.}
\label{tab:attack_acc_random_initial}
\begin{tabular}{@{\hspace{1cm}}ll cccccc}
\toprule
\multirow{2}{*}{\textbf{Network}} & \multirow{2}{*}{\textbf{Training}}  & \multirow{2}{*}{Clean} &
\multicolumn{4}{c}{\textbf{Accuracy under attack}}
\\\cmidrule{4-8}
 & {\textbf{Model}} &  & FWnucl 20$\,^{*}$ & FWnucl 20$\,^{+}$ & PGD 20 &  FGSM & CW \\
\midrule
\multicolumn{6}{l}{\textbf{MNIST}}\\\cmidrule{1-1}
\multirow{2}{*}{LeNET} & {Madry} & 98.38& {\bf 95.00} & {\bf 85.37} & 95.79 & 96.59 & 98.38 \\
& {ME-NET} & 99.24 & 97.02 & 52.63 & 74.88 & {\bf  46.18} & 99.00\\
\addlinespace
 \midrule
 \multirow{2}{*}{SmallCNN} & {Madry} & 99.12 & 98.03 & 92.35 & {\bf 95.77} & 97.95 & 99.12\\
& {ME-NET} & 99.42 & 89.04 & 71.75 & 76.84 & {\bf 54.09} & 91.44\\
\addlinespace
 \midrule
 \midrule
\multicolumn{6}{l}{\textbf{CIFAR-10}}\\\cmidrule{1-1}
\multirow{2}{*}{ResNet-18} & {Madry} & 81.25 & {\bf 34.64} & {\bf 0.45} & 49.95 & 55.91 & 78.61\\
& {ME-NET} & 92.09 & 29.66 & {\bf 4.01} & 4.99 & 44.80 & 79.57\\
\addlinespace
\midrule
\multirow{2}{*}{WideResNet} & {Madry} & 85.1 & {\bf 32.3} & {\bf 0.33} & 52.49 & 59.06 & 82.69\\
& {ME-NET} & 92.09 & 24.16 & {\bf 4.07} &  12.73 & 59.33 & 80.89 \\
\midrule
\multirow{2}{*}{ResNet-50} & {Madry} & 87.03 & {\bf 29.24} & {\bf 0.31} & 53.01 & 61.44 & 84.72 \\
& {ME-NET} &  92.09 & 28.49 & {\bf 4.09} &  9.14 & 58.51 & 83.21\\
\addlinespace
\bottomrule
\end{tabular}
\end{table}

\begin{figure}[htbp]
\subfloat[]{
\centering
\includegraphics[width=1.0\linewidth]{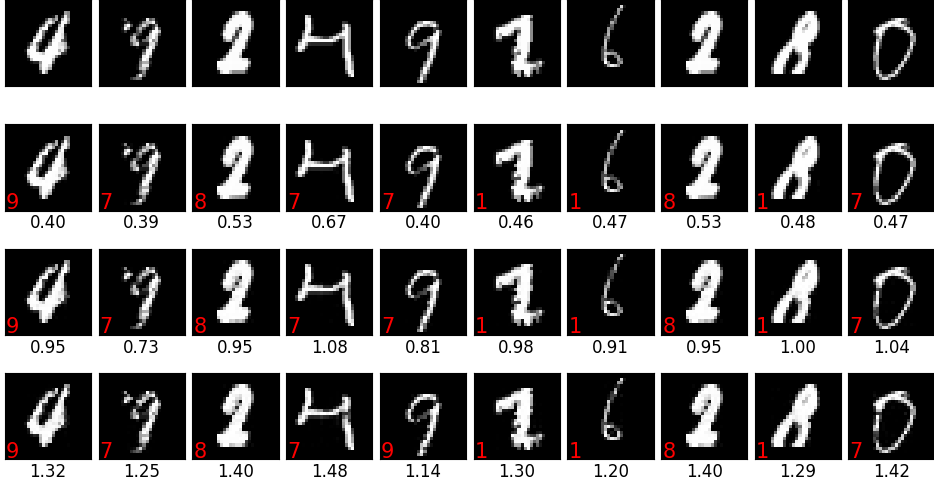}}%
\setlength{\abovecaptionskip}{2pt}
\caption{FWnucl adversarial examples for the MNIST dataset for different radii, (a) original images (first row), adversarial example generated by (b) FWnucl with $\epsilon_{S1}=1$ (second row), (c) FWnucl with $\epsilon_{S1}=3$ (third row) and (d) FWnucl with $\epsilon_{S1}=5$ (third row). The fooling label is shown on the image.}
\label{mnist-radii-images}
\end{figure}

\begin{figure}[htbp]
\subfloat{
\centering
\includegraphics[width=1.0\linewidth]{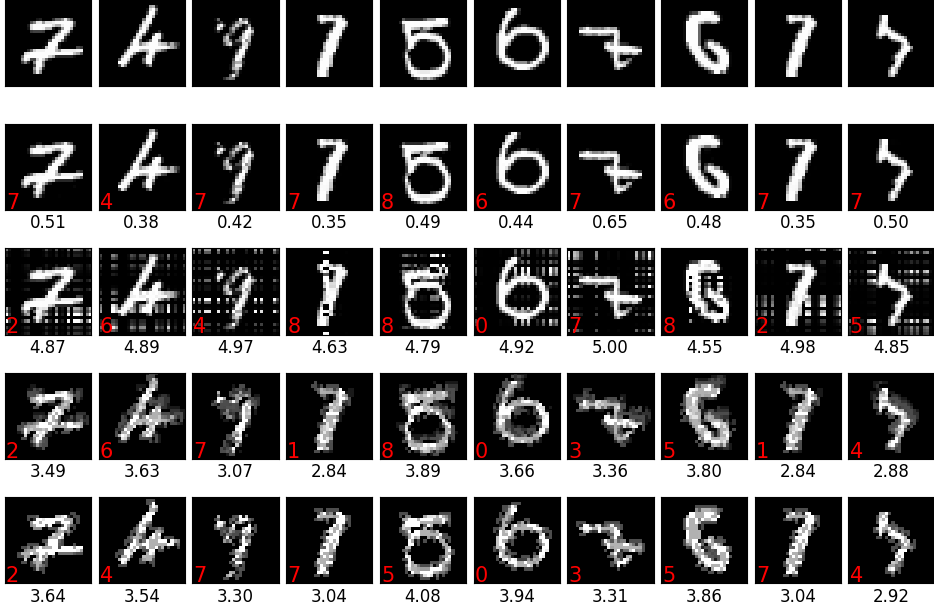}}%
\setlength{\abovecaptionskip}{2pt}
\caption{Adversarial examples on MNIST dataset  against the Madry defense (a) original images (first row), adversarial examples generated by (b) FWnucl (second row), (c) PGDnucl (third row), (d) PGD (fourth row) and (e) FGSM (fifth row); the fooling label is shown on bottom-right of each image, and the $l_2$ norm of adversarial noise is written below each image.}
\label{mnist-cifar-images}
\end{figure}

\begin{figure}[h!]
\centering
\subfloat[stole]{{\includegraphics[scale=0.32]{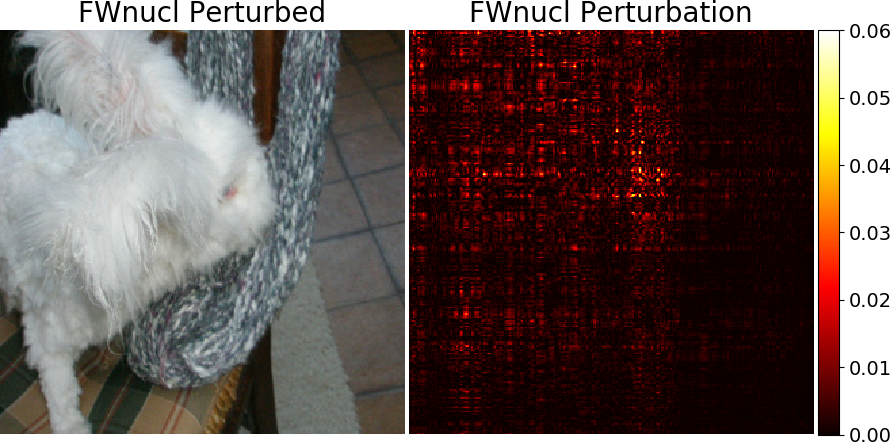}}}\hfill
\subfloat[stole]{{\includegraphics[scale=0.32]{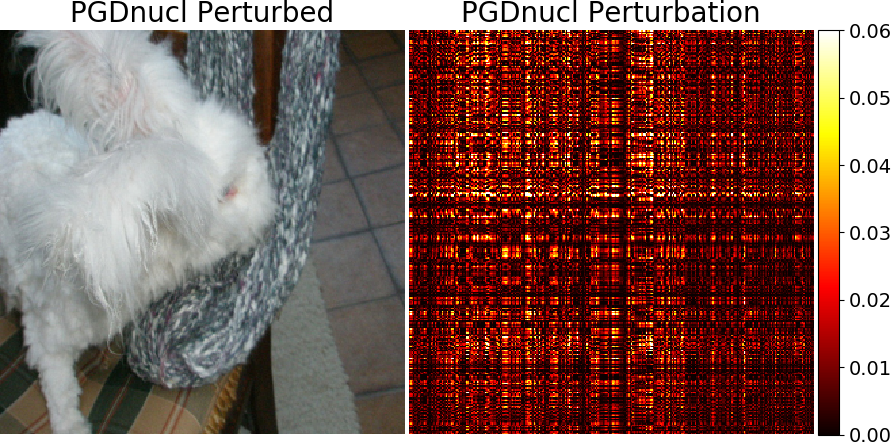}}}\\
\subfloat[stole]{{\includegraphics[scale=0.32]{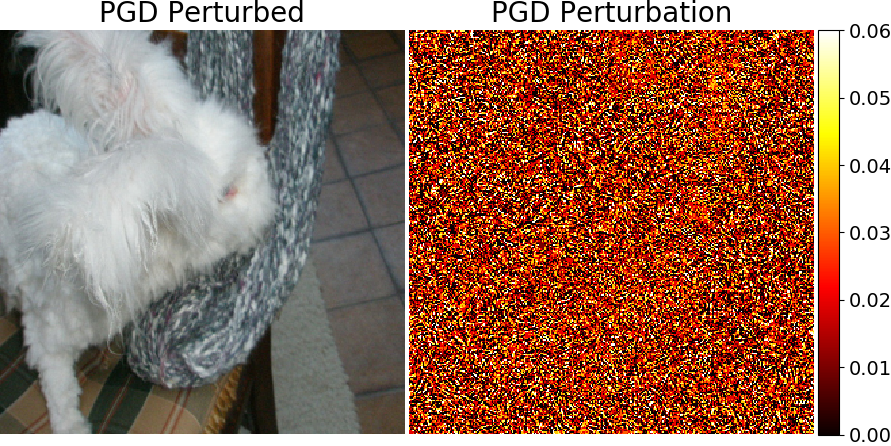}}}\hfill
\subfloat[stole]{{\includegraphics[scale=0.32]{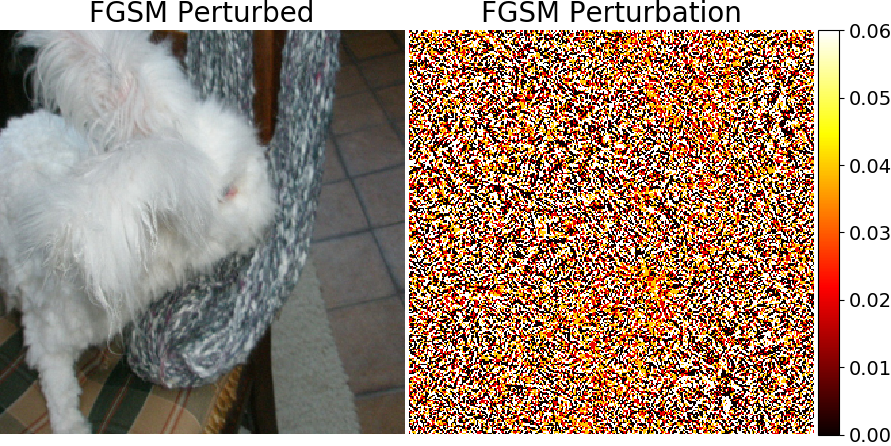}}}
\end{figure}

\begin{figure}[h!]
\centering
\subfloat[FWnucl]{{\includegraphics[scale=0.42]{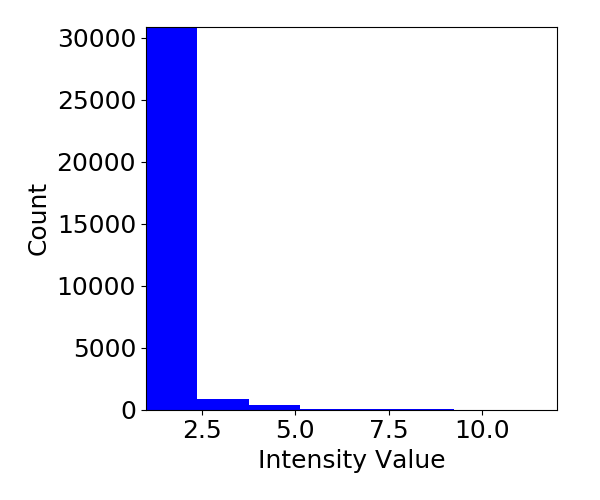}}}\hfill
\subfloat[PGDnucl]{{\includegraphics[scale=0.42]{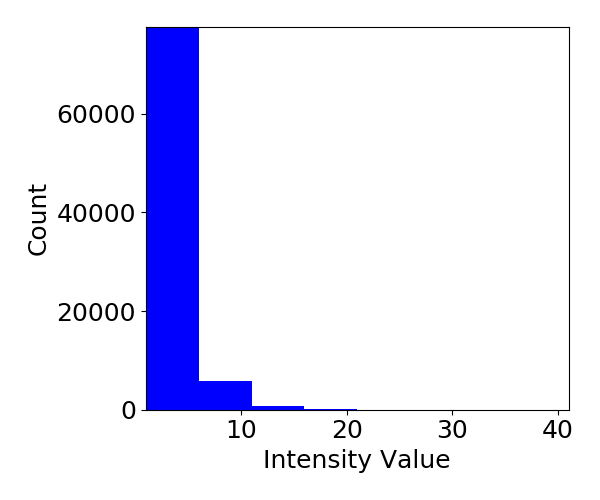}}}\\
\subfloat[PGD]{{\includegraphics[scale=0.42]{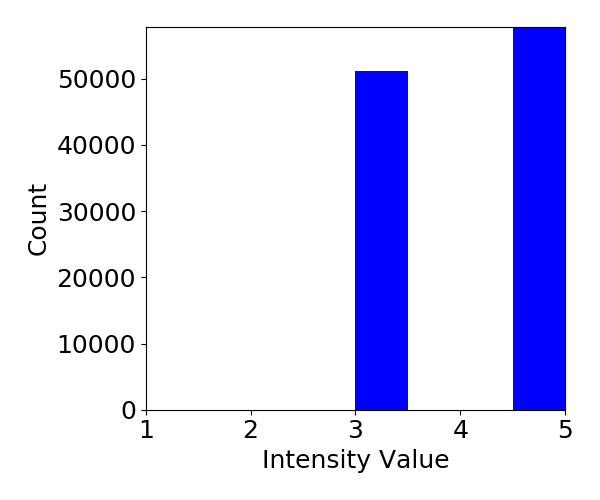}}}\hfill
\subfloat[FGSM]{{\includegraphics[scale=0.42]{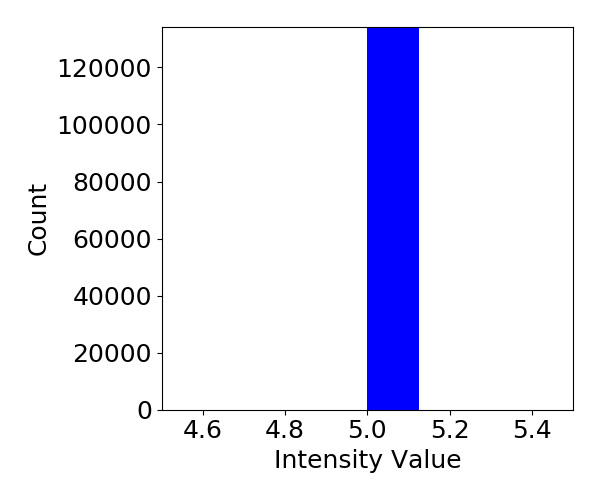}}}
\caption{ResNet-50 adversarial examples with the histogram of non-zero pixel intensities corresponding to the attack perturbations for the {\bf Angora} original image label.}
\end{figure}

\begin{figure}[h!]
\centering
\subfloat[tray]{{\includegraphics[scale=0.32]{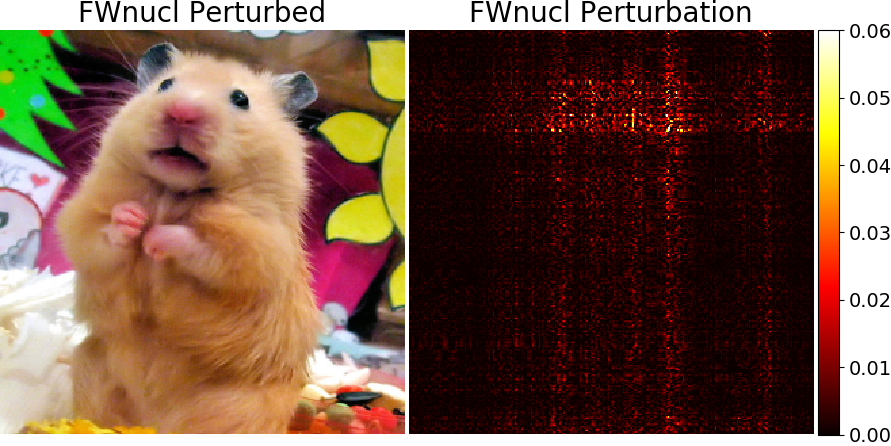}}}\hfill
\subfloat[hamster]{{\includegraphics[scale=0.32]{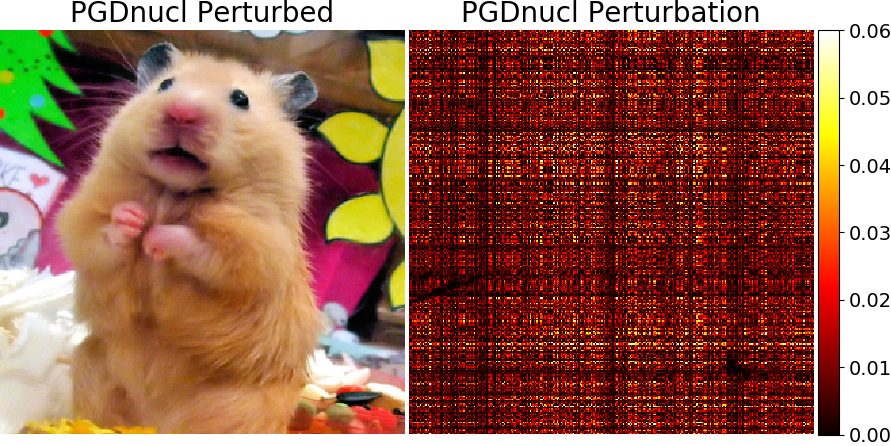}}}\\
\subfloat[Irish setter]{{\includegraphics[scale=0.32]{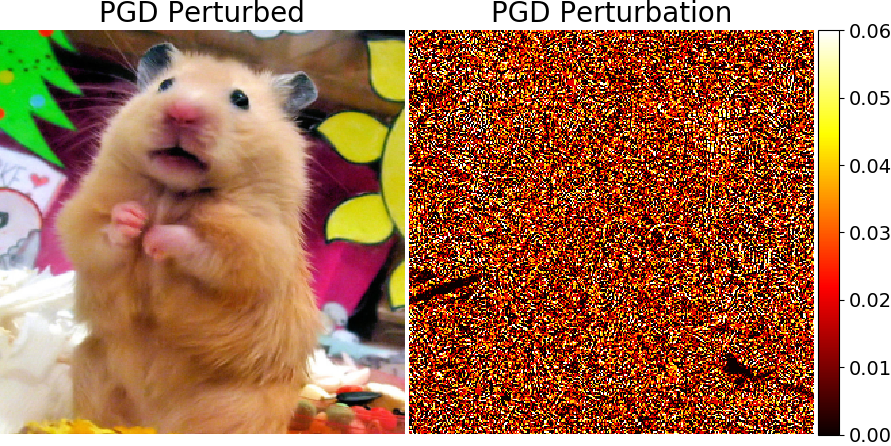}}}\hfill
\subfloat[Persian cat]{{\includegraphics[scale=0.32]{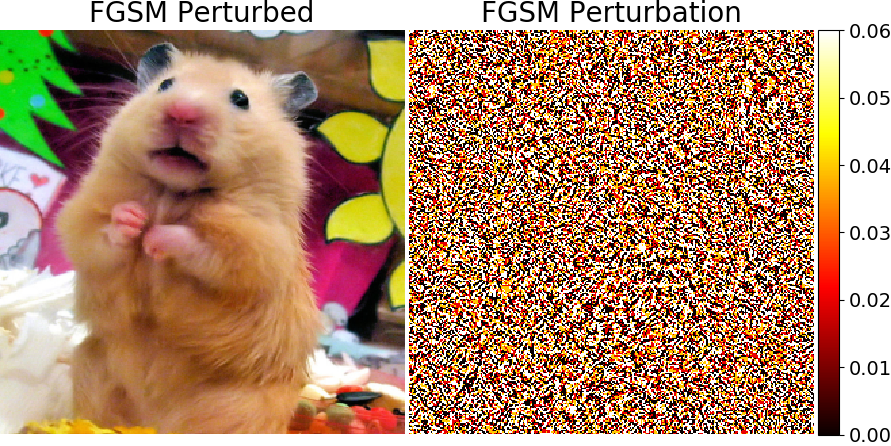}}}
\caption{ResNt-50 adversarial examples with the heat map of pixel intensities corresponding to the attack perturbations for the original image with the label {\bf hamster}.}
\end{figure}

\begin{figure}[h!]
\centering
\subfloat[clog]{{\includegraphics[scale=0.32]{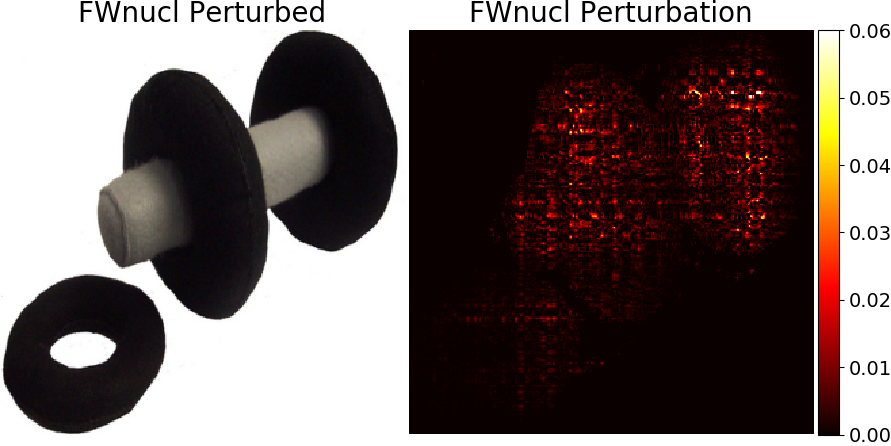}}}\hfill
\subfloat[ski mask]{{\includegraphics[scale=0.32]{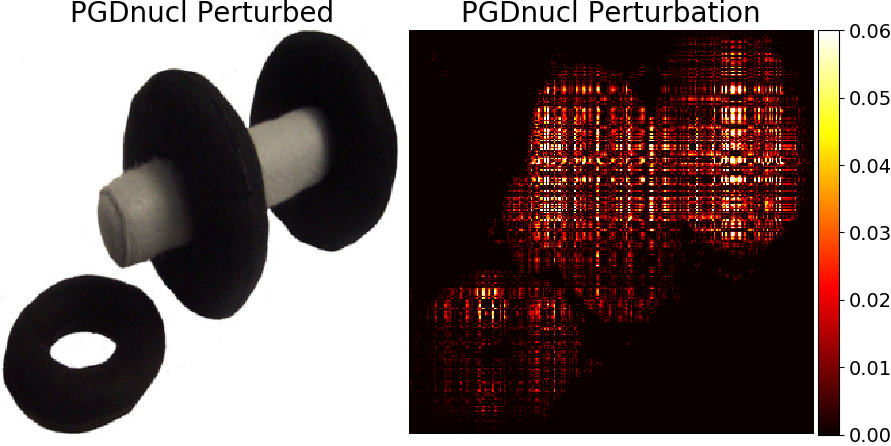}}}\\
\subfloat[ski mask]{{\includegraphics[scale=0.32]{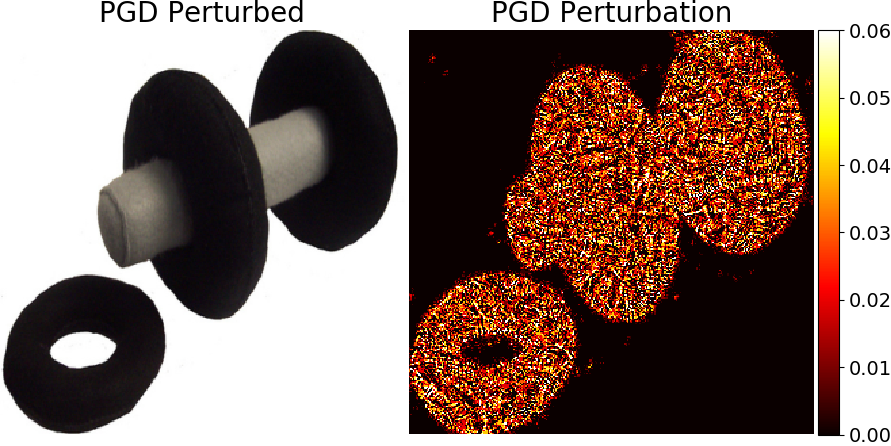}}}\hfill
\subfloat[dumbbell]{{\includegraphics[scale=0.32]{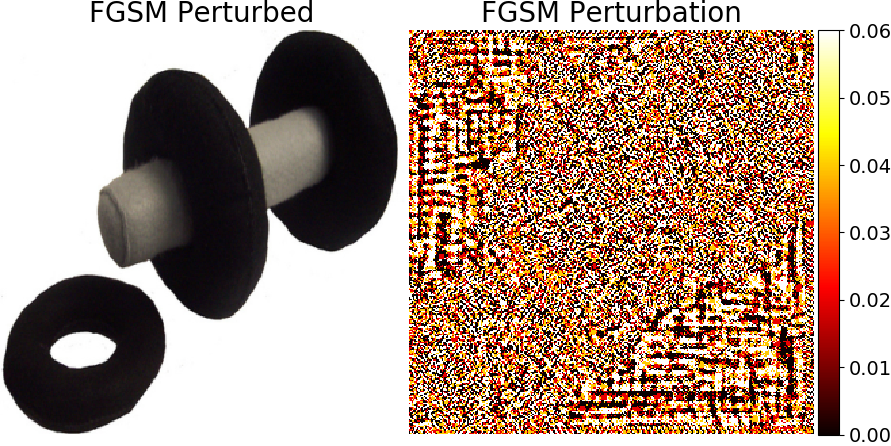}}}
\caption{ResNet-50 adversarial examples with the heat map of pixel intensities corresponding to the attack perturbations for the original image with the label {\bf dumbbell}.}
\end{figure}

\begin{figure}[h!]
\centering
\subfloat[patio]{{\includegraphics[scale=0.32]{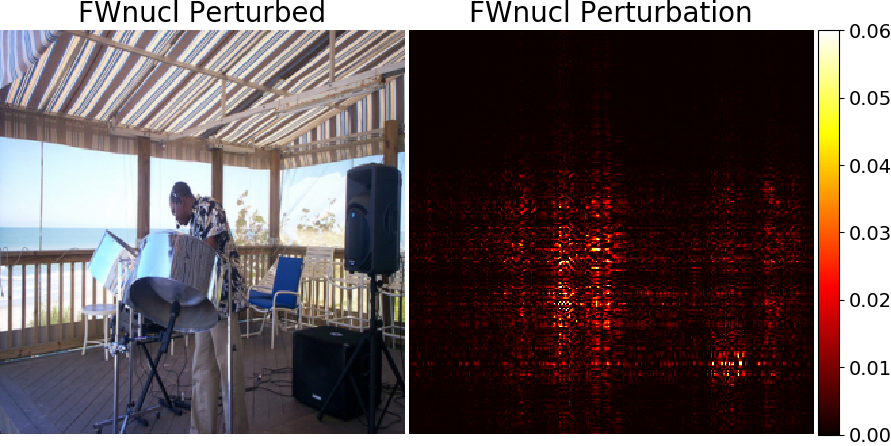}}}\hfill
\subfloat[steel drum]{{\includegraphics[scale=0.32]{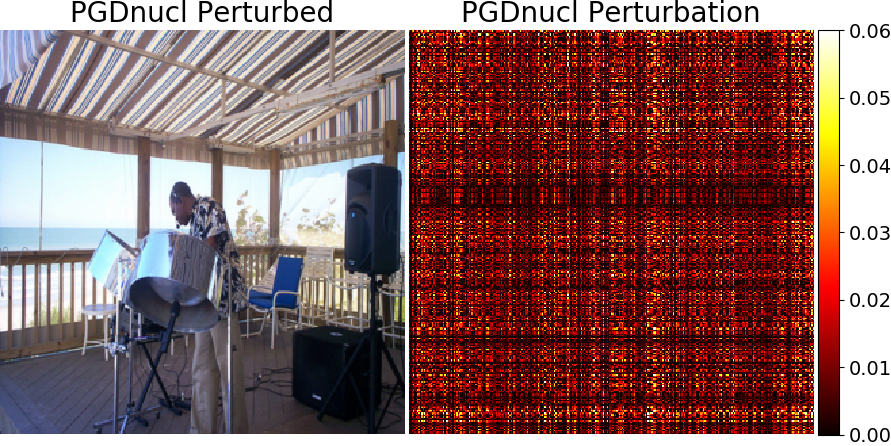}}}\\
\subfloat[patio]{{\includegraphics[scale=0.32]{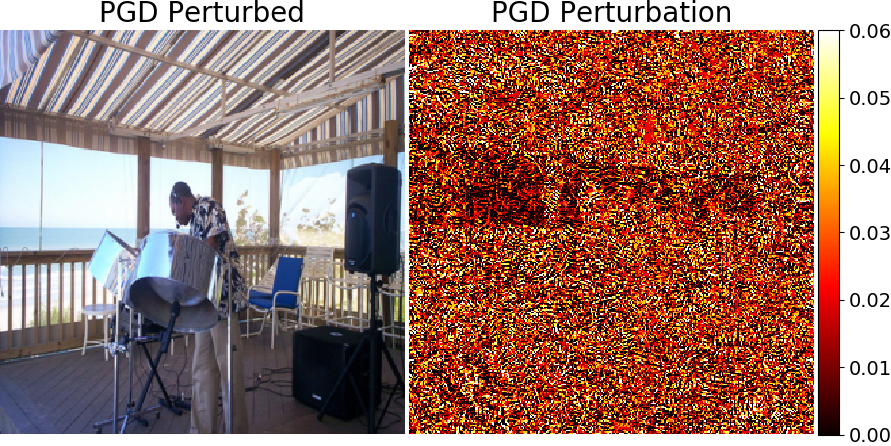}}}\hfill
\subfloat[folding chair]{{\includegraphics[scale=0.32]{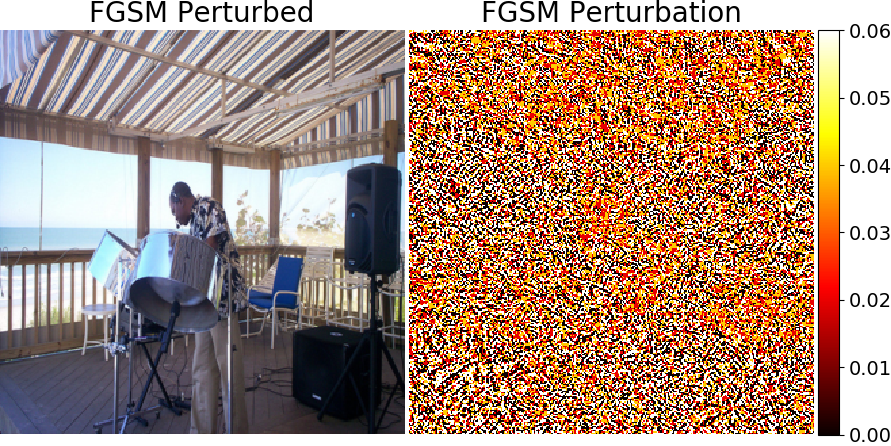}}}\caption{ResNet-50 adversarial examples with the heat map of pixel intensities corresponding to the attack perturbations for the original image with the label {\bf steel drum}.}
\end{figure}

\begin{figure}[h!]
\centering
\subfloat[beach wagon]{{\includegraphics[scale=0.32]{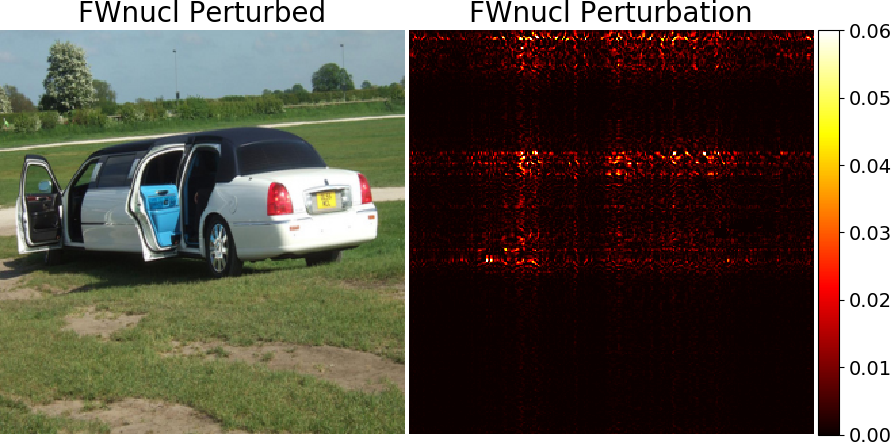}}}\hfill
\subfloat[beach wagon]{{\includegraphics[scale=0.32]{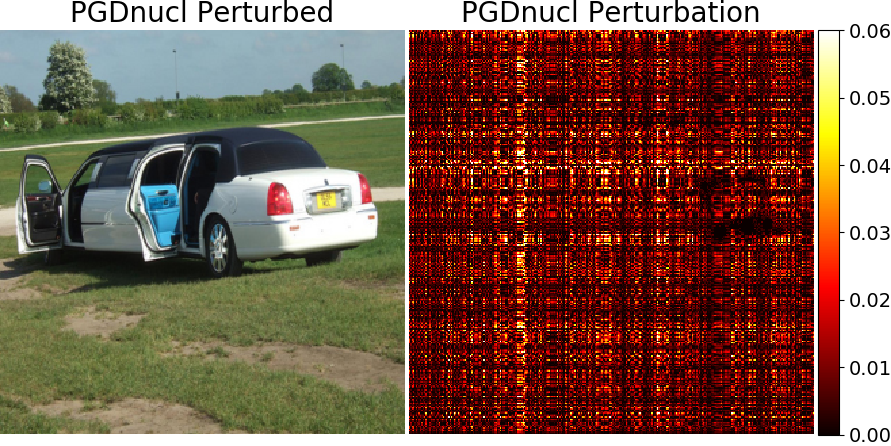}}}\\
\subfloat[black stork]{{\includegraphics[scale=0.32]{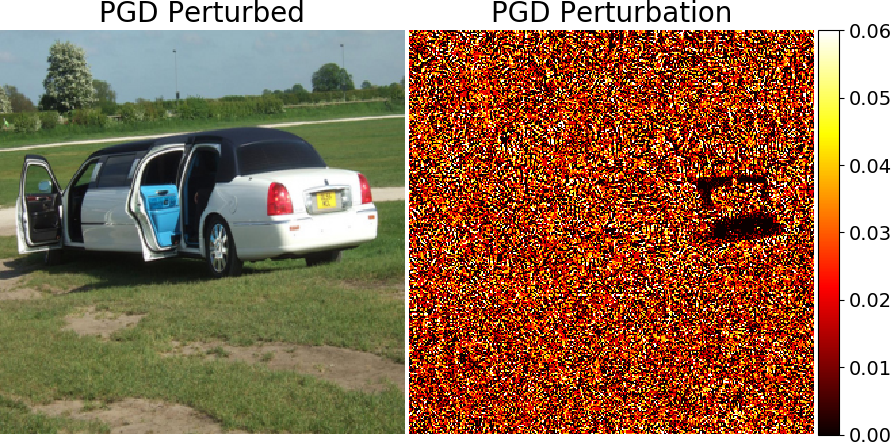}}}\hfill
\subfloat[beach wagon]{{\includegraphics[scale=0.32]{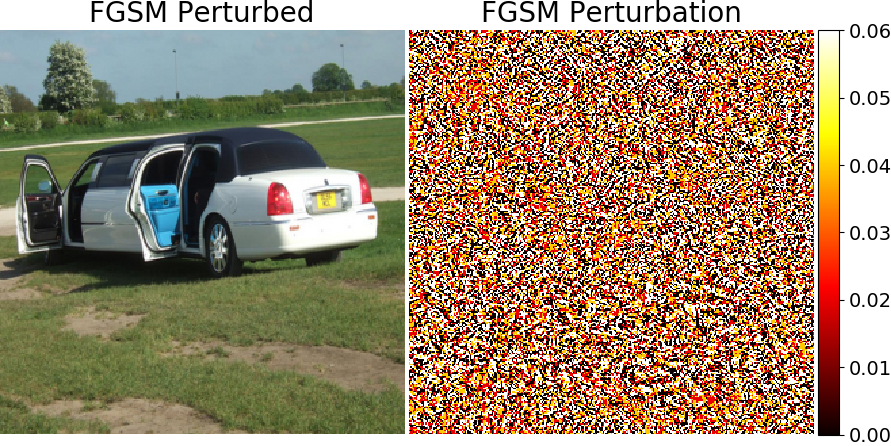}}}\caption{ResNet-50 adversarial examples with the heat map of pixel intensities corresponding to the attack perturbations for the original image with the label {\bf limousine}.}
\end{figure}

\begin{figure}[h!]
\centering
\subfloat[barometer]{{\includegraphics[scale=0.32]{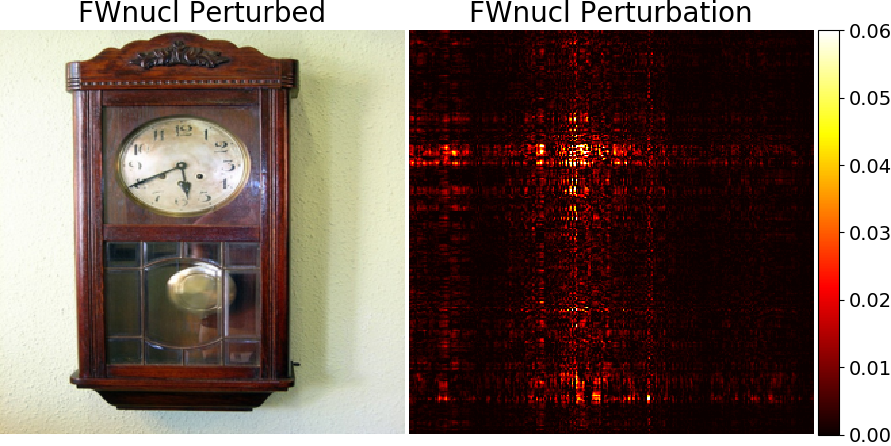}}}\hfill
\subfloat[wall clock]{{\includegraphics[scale=0.32]{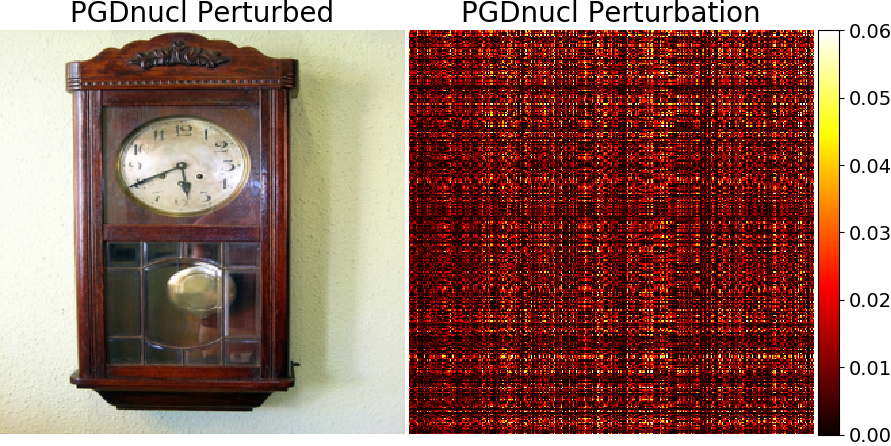}}}\\
\subfloat[wardrobe]{{\includegraphics[scale=0.32]{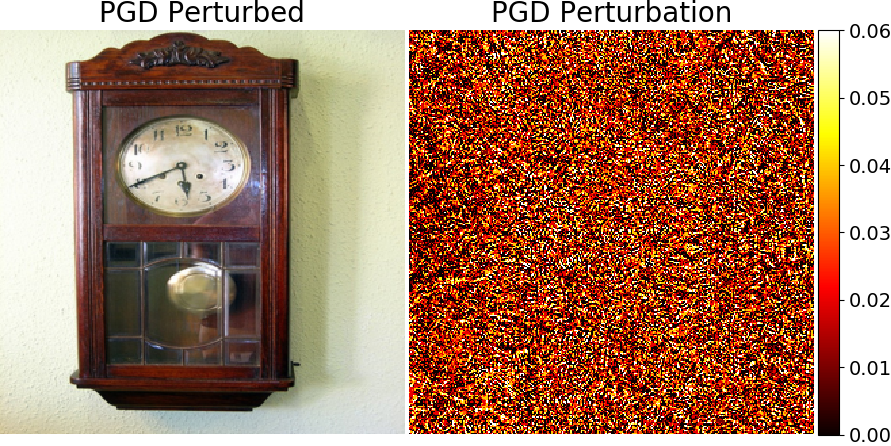}}}\hfill
\subfloat[wardrobe]{{\includegraphics[scale=0.32]{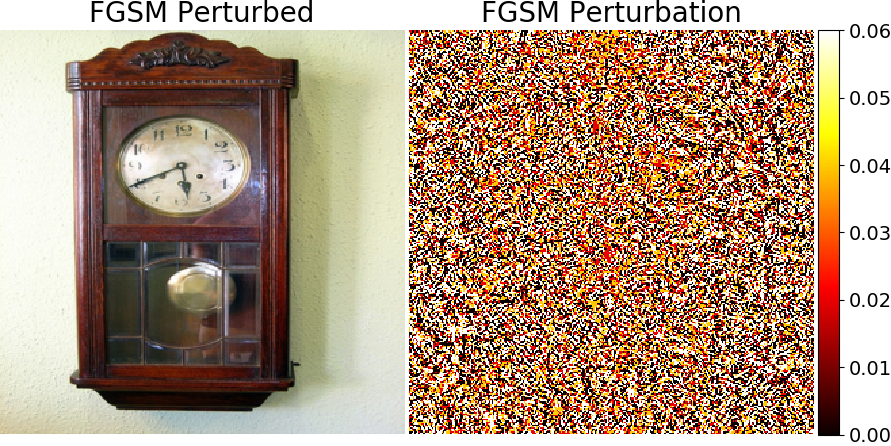}}}
\caption{ResNet-50 adversarial examples with the heat map of pixel intensities corresponding to the attack perturbations for the original image with the label {\bf wall clock}.}
\end{figure}

\begin{figure}[h!]
\centering
\subfloat[flagpole]{{\includegraphics[scale=0.32]{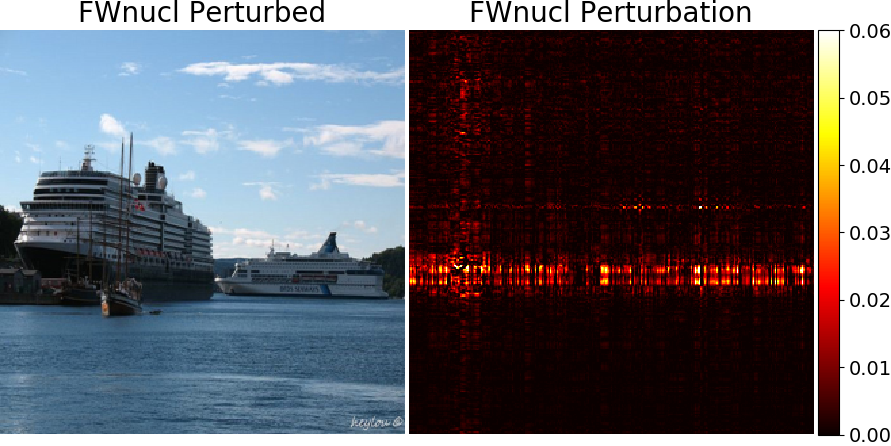}}}\hfill
\subfloat[liner]{{\includegraphics[scale=0.32]{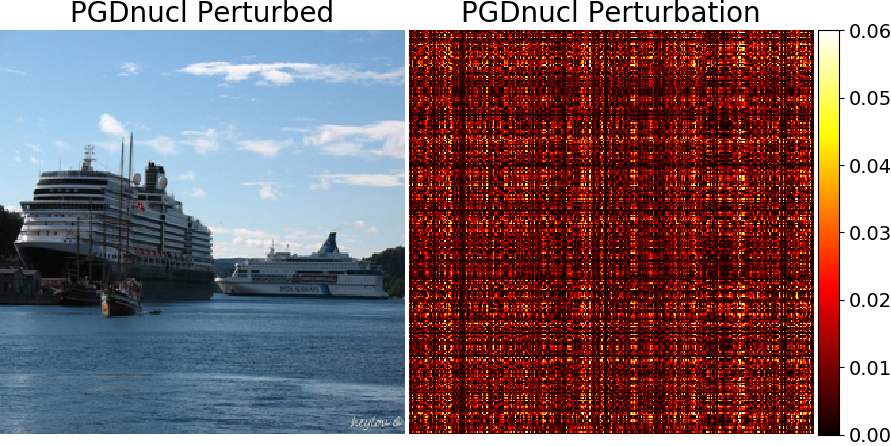}}}\\
\subfloat[bath towel]{{\includegraphics[scale=0.32]{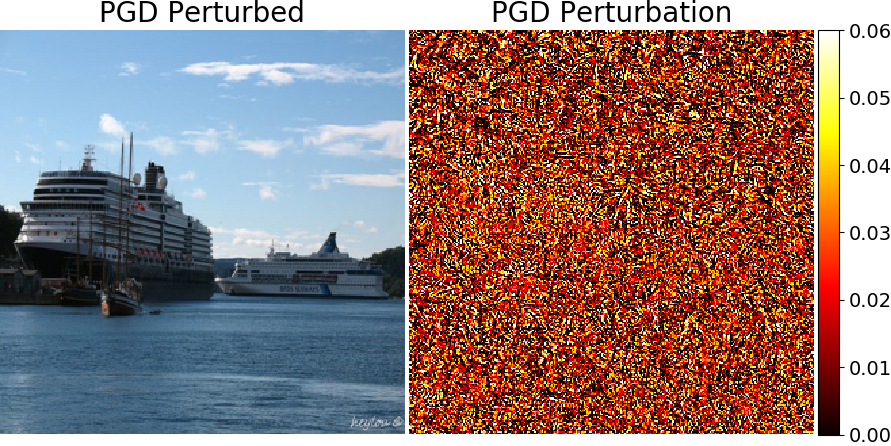}}}\hfill
\subfloat[fountain 3.34]{{\includegraphics[scale=0.32]{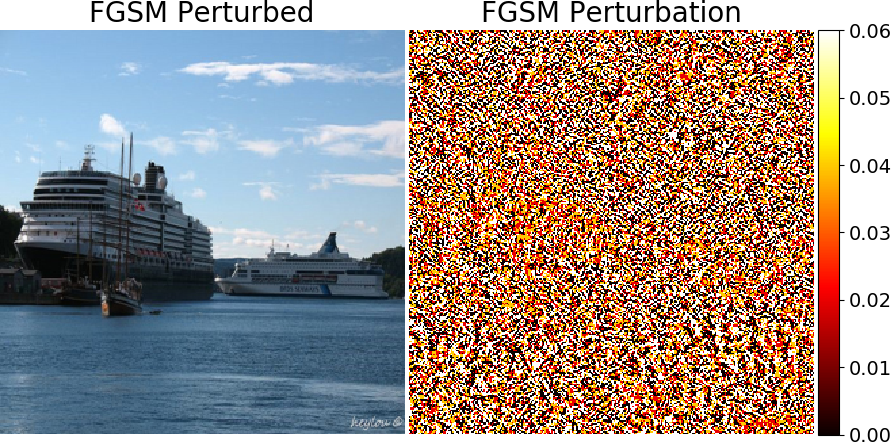}}}
\caption{ResNet-50 adversarial examples with the heat map of pixel intensities corresponding to the attack perturbations for the original image with the label {\bf liner}.}
\end{figure}

\begin{figure}[h!]
\centering
\subfloat[car wheel]{{\includegraphics[scale=0.32]{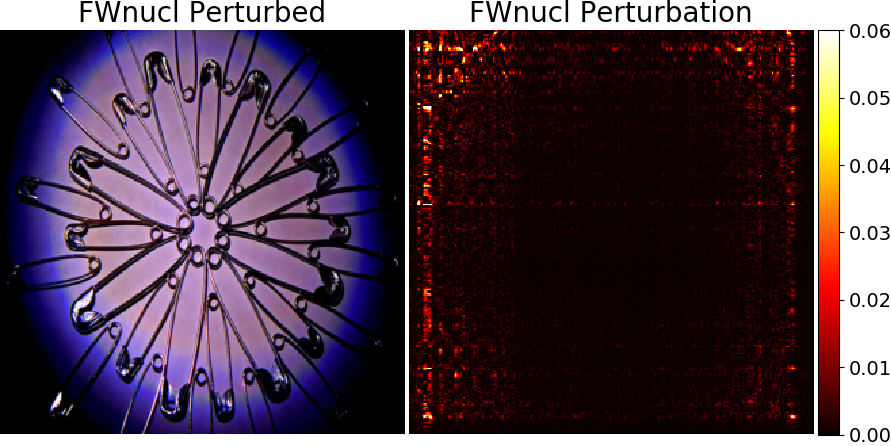}}}\hfill
\subfloat[dome]{{\includegraphics[scale=0.32]{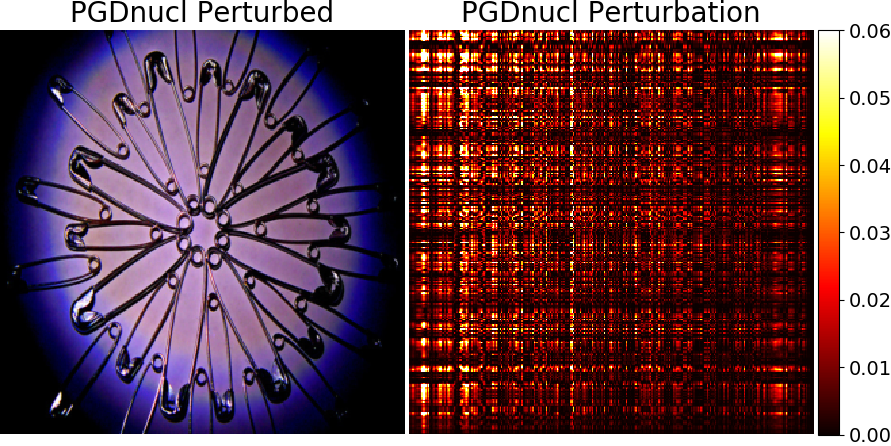}}}\\
\subfloat[dome]{{\includegraphics[scale=0.32]{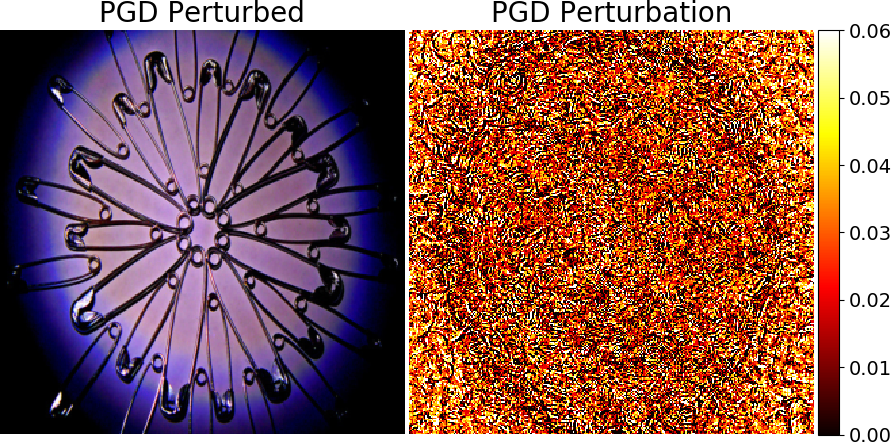}}}\hfill
\subfloat[wall clock]{{\includegraphics[scale=0.32]{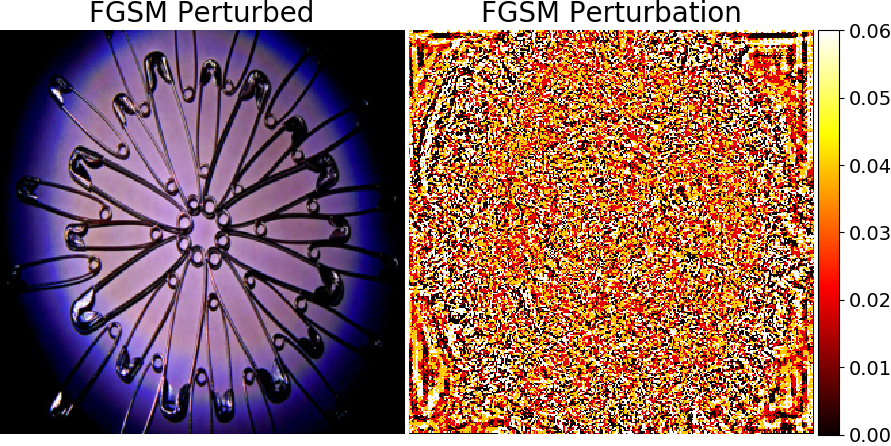}}}
\caption{DenseNet121 adversarial examples with the heat map of pixel intensities corresponding to the attack perturbations for the original image with the label {\bf safety pin}.}
\end{figure}


\begin{figure}[h!]
\centering
\subfloat[barrow]{{\includegraphics[scale=0.32]{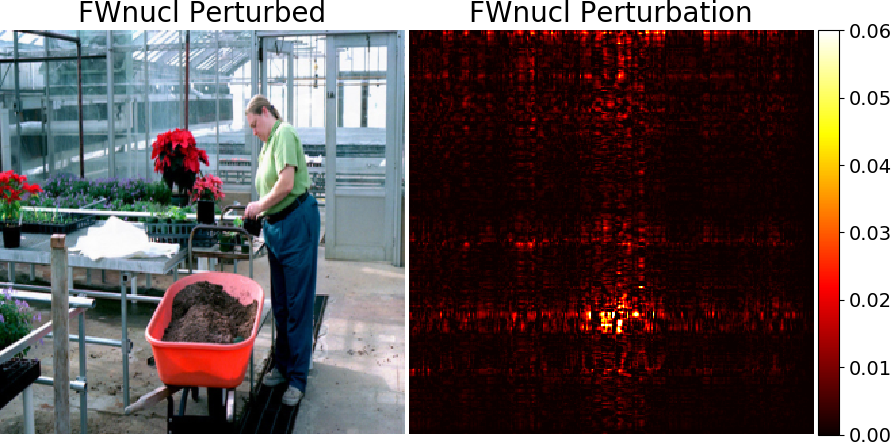}}}\hfill
\subfloat[barrow]{{\includegraphics[scale=0.32]{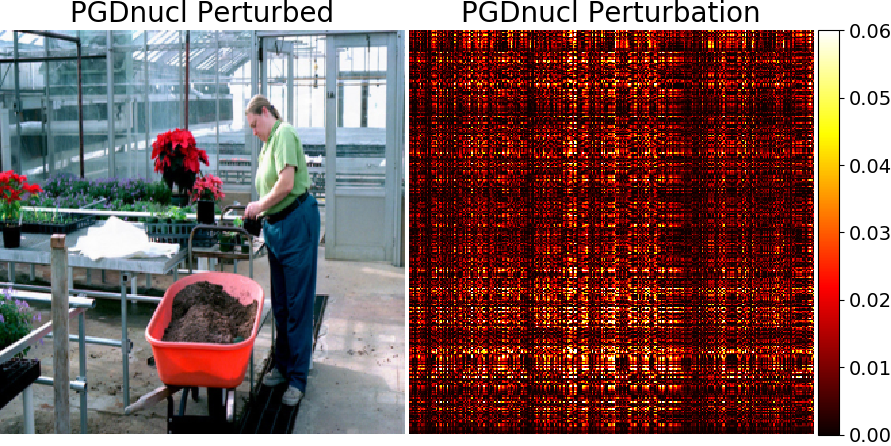}}}\\
\subfloat[barrow]{{\includegraphics[scale=0.32]{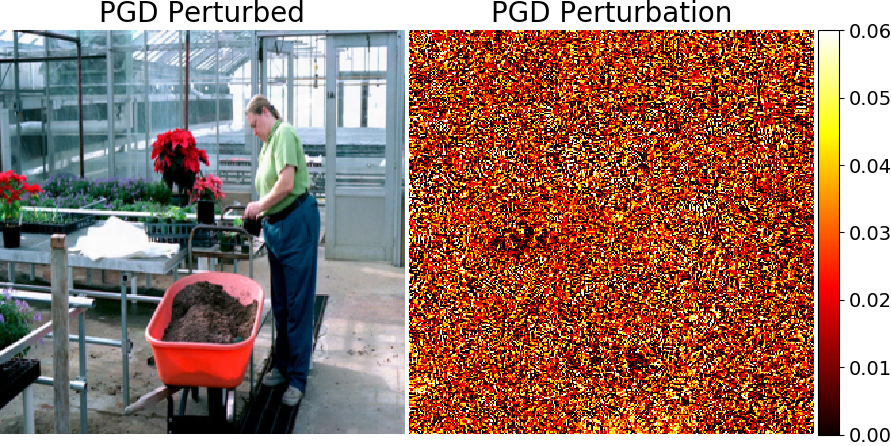}}}\hfill
\subfloat[barrow]{{\includegraphics[scale=0.32]{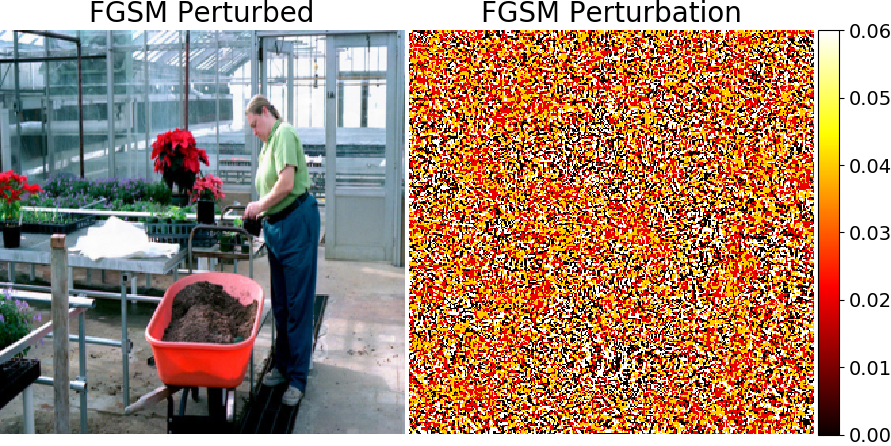}}}
\caption{DenseNet121 adversarial examples with the heat map of pixel intensities corresponding to the attack perturbations for the original image with the label {\bf greenhouse}.}
\end{figure}

\begin{figure}[h!]
\centering
\subfloat[parachute]{{\includegraphics[scale=0.32]{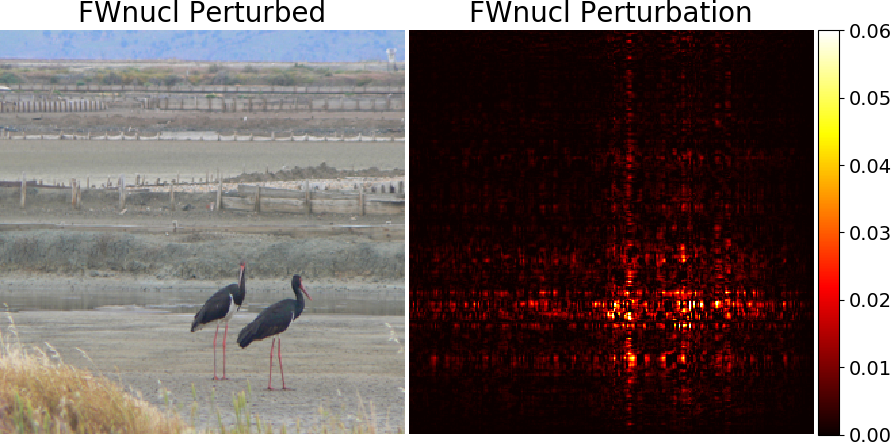}}}\hfill
\subfloat[black stork]{{\includegraphics[scale=0.32]{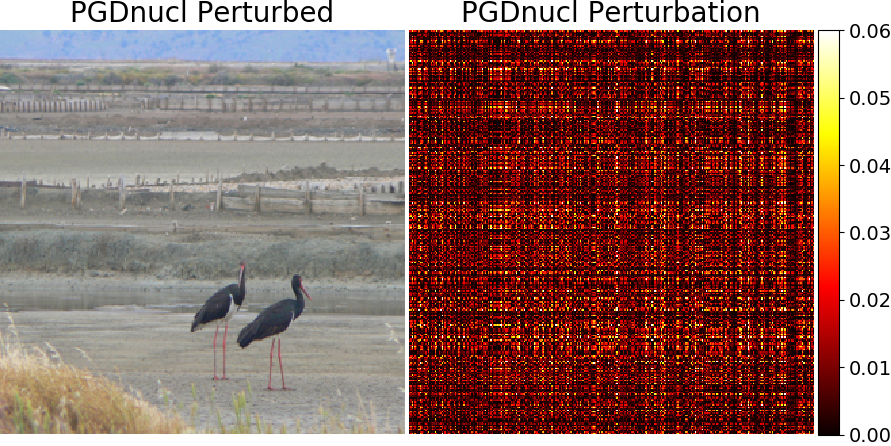}}}\\
\subfloat[maze]{{\includegraphics[scale=0.32]{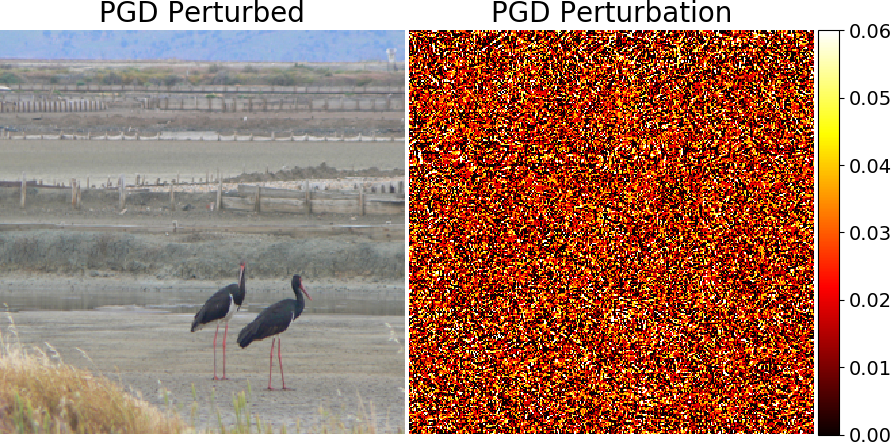}}}\hfill
\subfloat[crane]{{\includegraphics[scale=0.32]{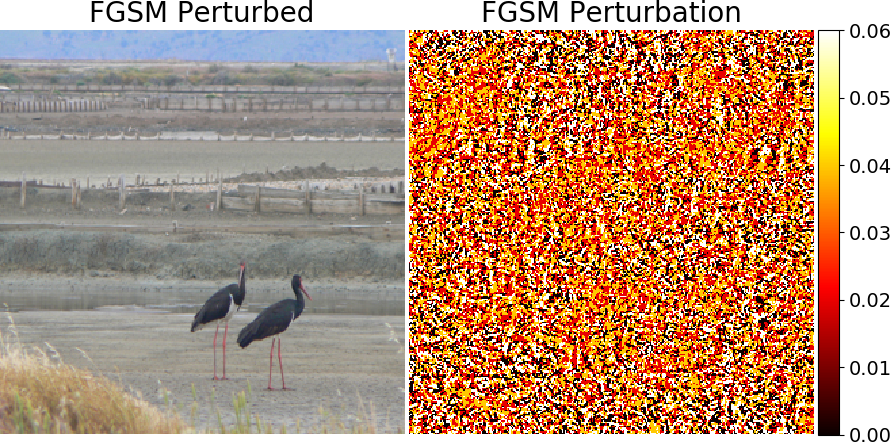}}}
\caption{DenseNet121 adversarial examples with the heat map of pixel intensities corresponding to the attack perturbations for the original image with the label {\bf black stork}.}
\end{figure}

\begin{figure}[h!]
\centering
\subfloat[vacuum]{{\includegraphics[scale=0.32]{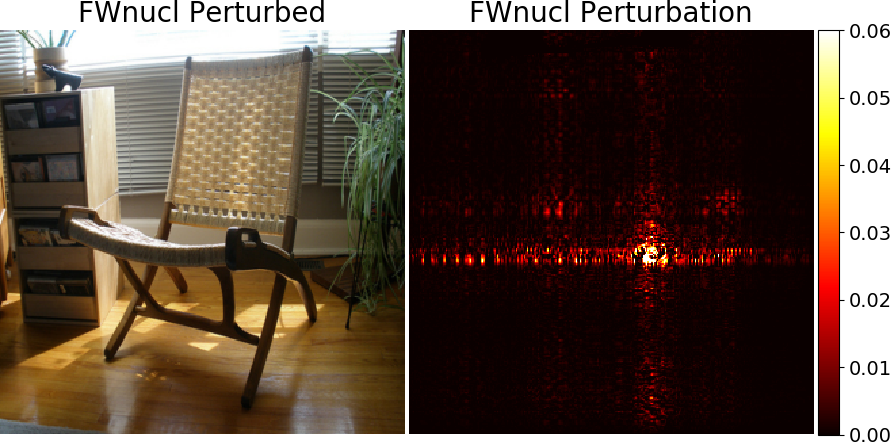}}}\hfill
\subfloat[rocking chair]{{\includegraphics[scale=0.32]{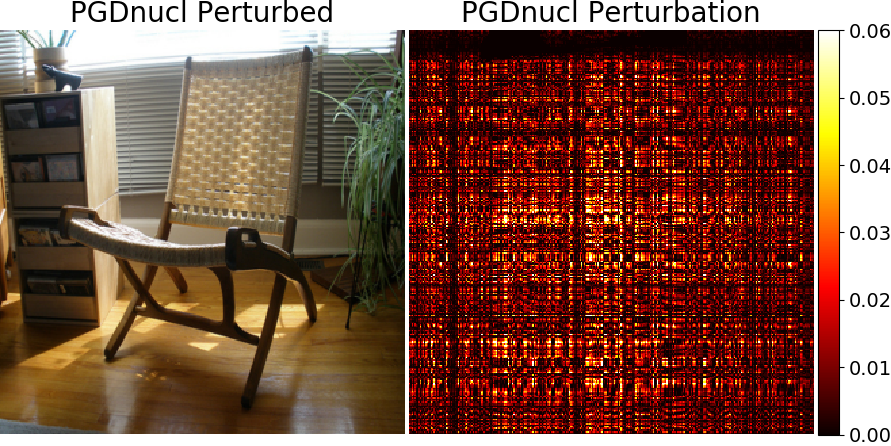}}}\\
\subfloat[rocking chair]{{\includegraphics[scale=0.32]{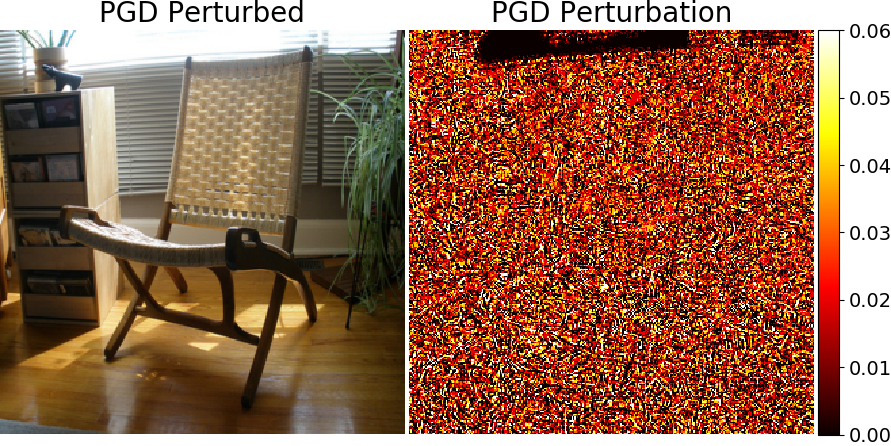}}}\hfill
\subfloat[rocking chair]{{\includegraphics[scale=0.32]{plots/Supplementaries/DenseNet121/images/black_storkfgsm.png}}}\caption{DenseNet121 adversarial examples with the heat map of pixel intensities corresponding to the attack perturbations for the original image with the label {\bf folding chair}.}
\end{figure}

\begin{figure}[h!]
\centering
\subfloat[television]{{\includegraphics[scale=0.32]{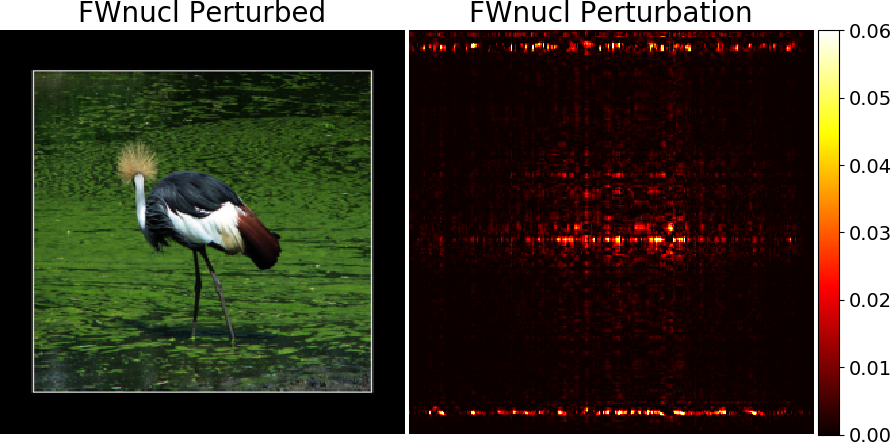}}}\hfill
\subfloat[white stork]{{\includegraphics[scale=0.32]{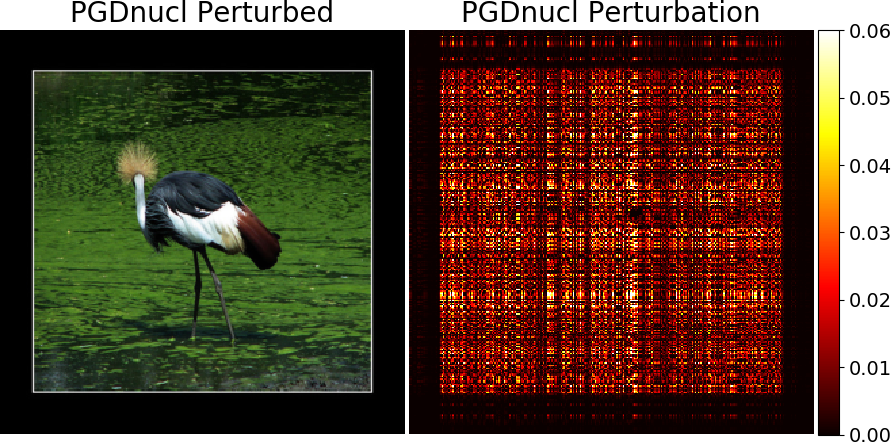}}}\\
\subfloat[monitor]{{\includegraphics[scale=0.32]{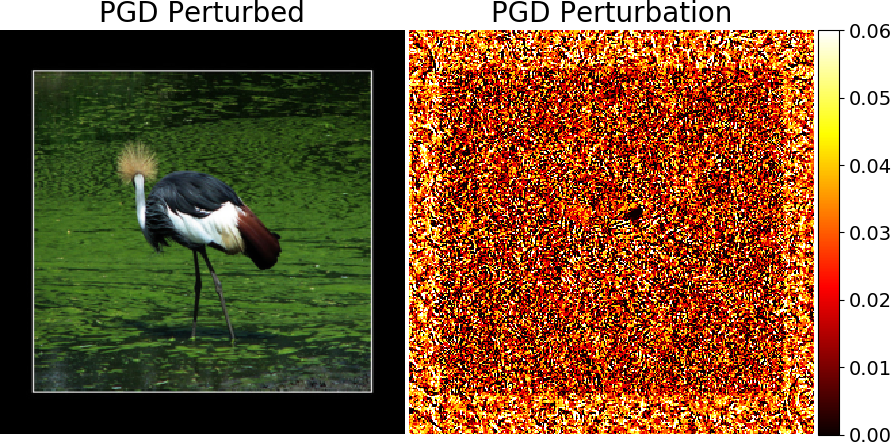}}}\hfill
\subfloat[white stork]{{\includegraphics[scale=0.32]{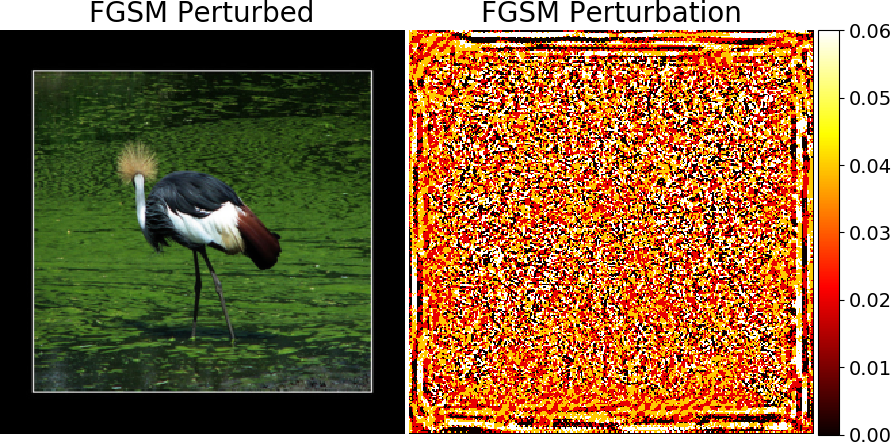}}}
\caption{DenseNet121 adversarial examples with the heat map of pixel intensities corresponding to the attack perturbations for the original image with the label {\bf crane}.}
\end{figure}

\begin{figure}[h!]
\centering
\subfloat[pickup truck]{{\includegraphics[scale=0.32]{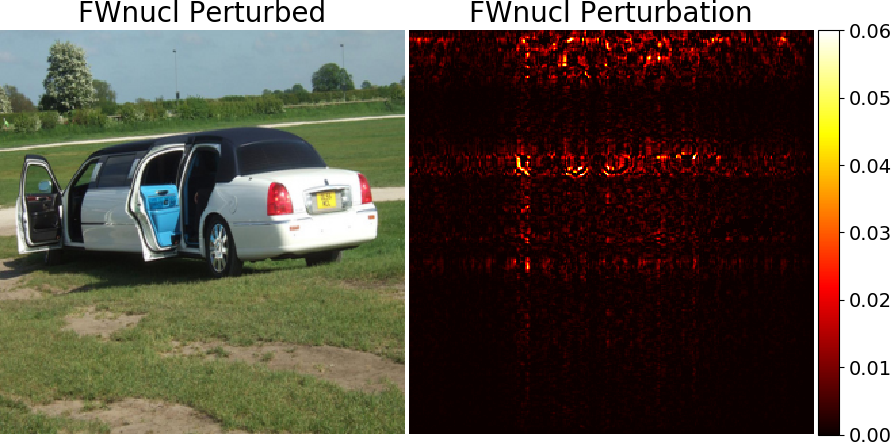}}}\hfill
\subfloat[pickup truck]{{\includegraphics[scale=0.32]{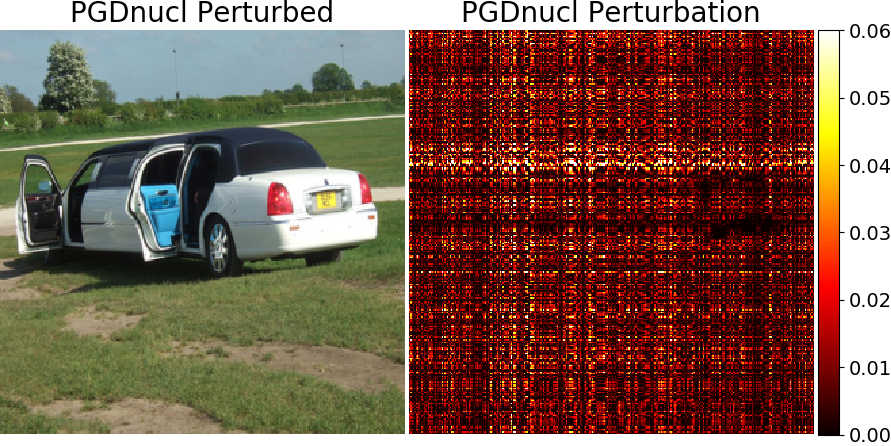}}}\\
\subfloat[racing car]{{\includegraphics[scale=0.32]{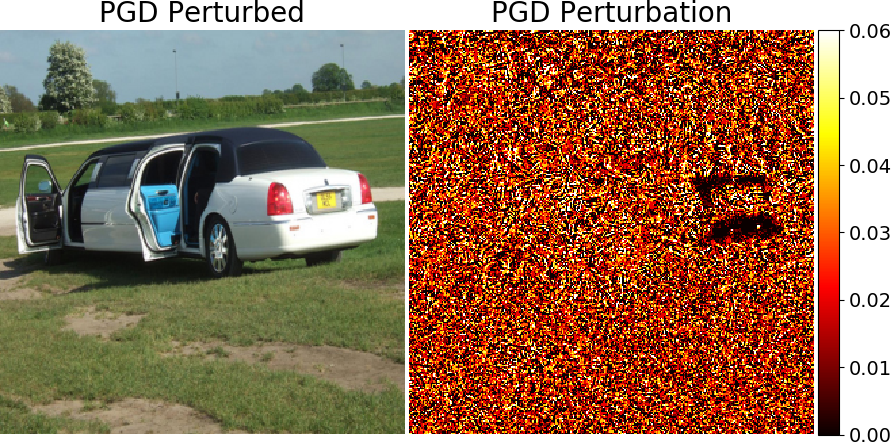}}}\hfill
\subfloat[racing car]{{\includegraphics[scale=0.32]{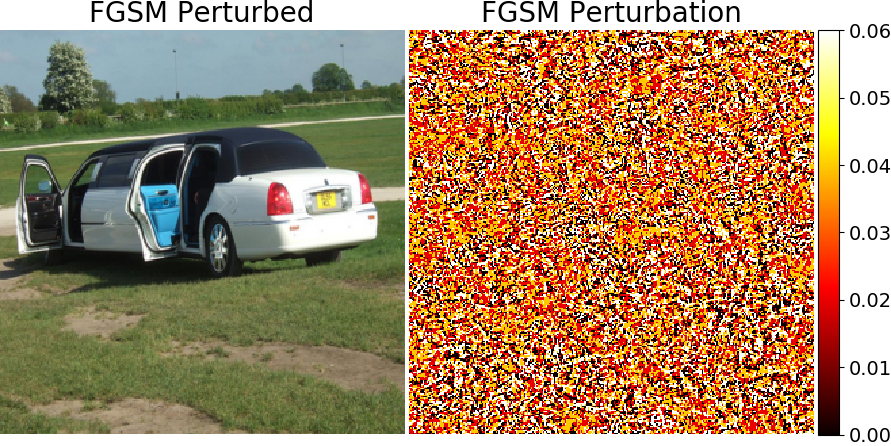}}}
\caption{DenseNet121 adversarial examples with the heat map of pixel intensities corresponding to the attack perturbations for the original image with the label {\bf limousine}.}
\end{figure}

\begin{figure}[h!]
\centering
\subfloat[vault]{{\includegraphics[scale=0.32]{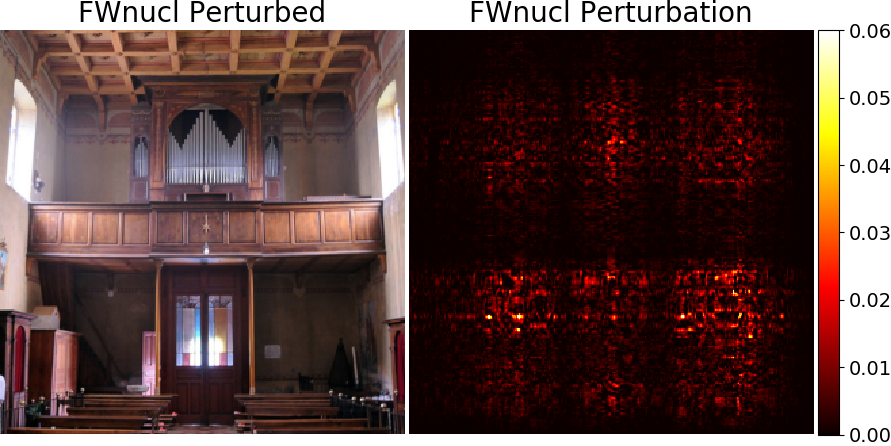}}}\hfill
\subfloat[vault]{{\includegraphics[scale=0.32]{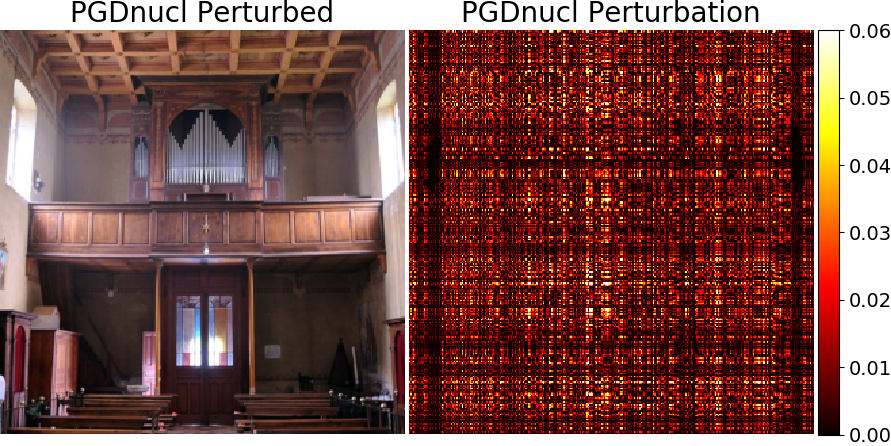}}}\\
\subfloat[vault]{{\includegraphics[scale=0.32]{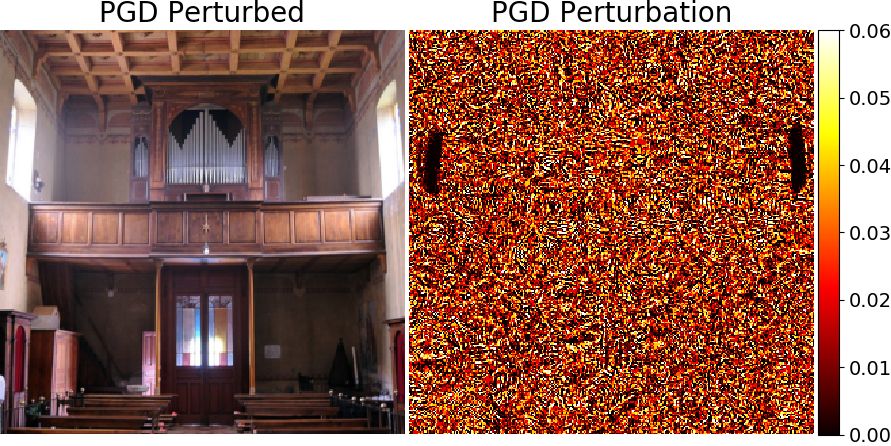}}}\hfill
\subfloat[vault]{{\includegraphics[scale=0.32]{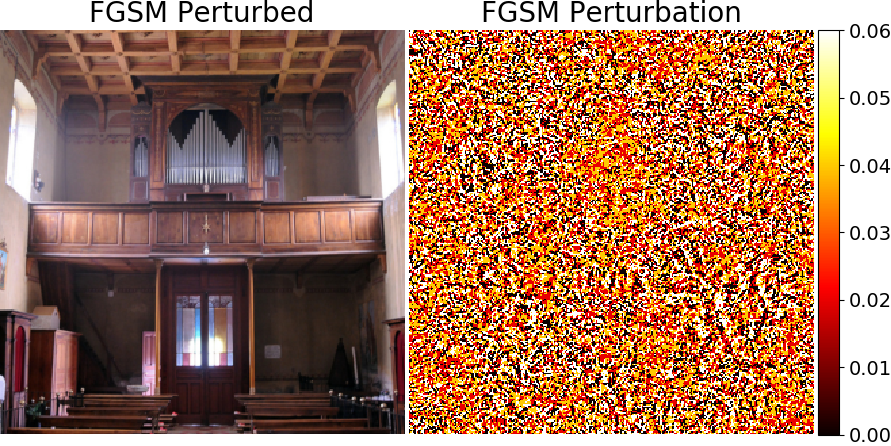}}}
\caption{DenseNet121 adversarial examples with the heat map of pixel intensities corresponding to the attack perturbations for the original image with the label {\bf organ}.}
\end{figure}

\begin{figure}[h!]
\centering
\subfloat[packet]{{\includegraphics[scale=0.32]{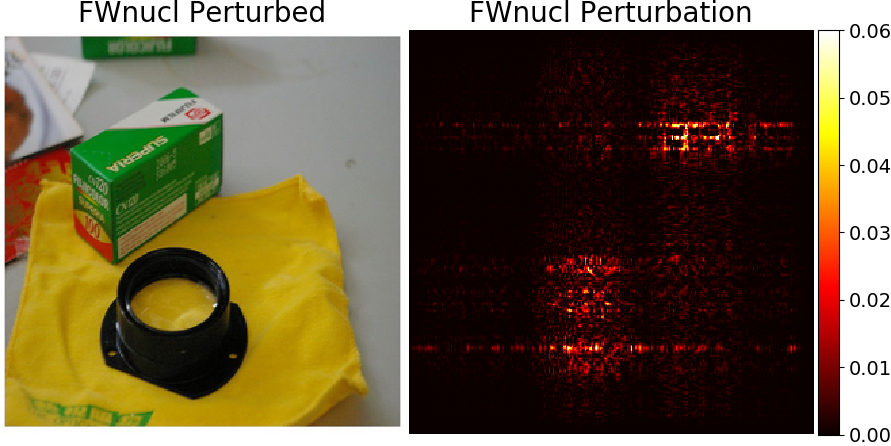}}}\hfill
\subfloat[packet]{{\includegraphics[scale=0.32]{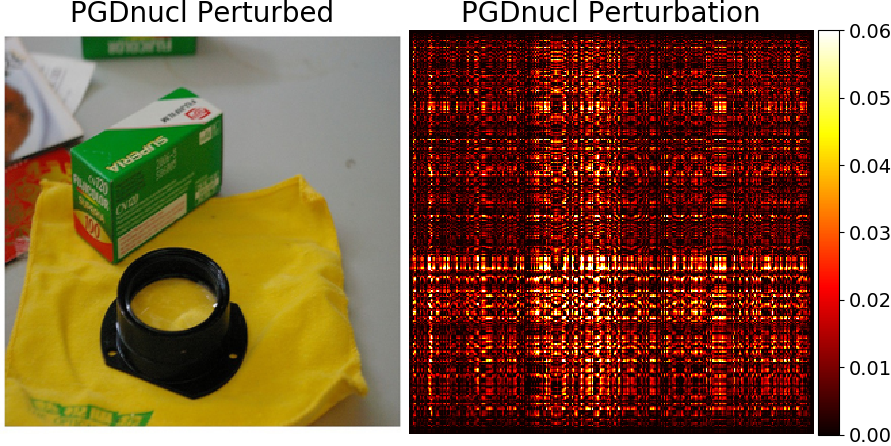}}}\\
\subfloat[plastic bag]{{\includegraphics[scale=0.32]{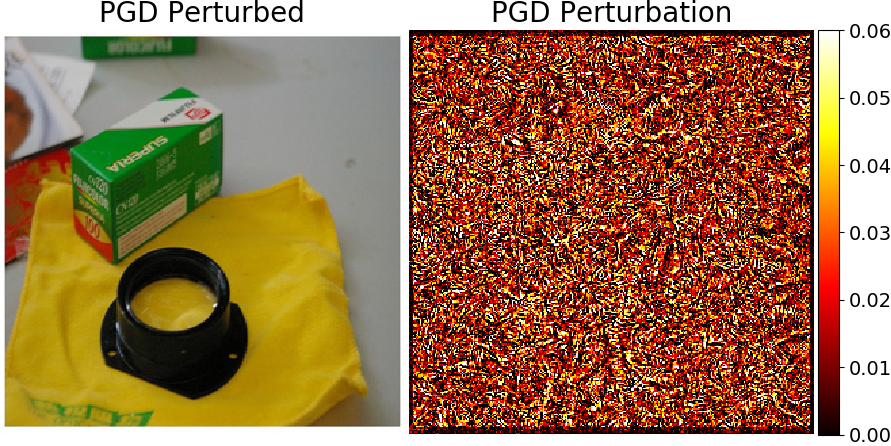}}}\hfill
\subfloat[bucket]{{\includegraphics[scale=0.32]{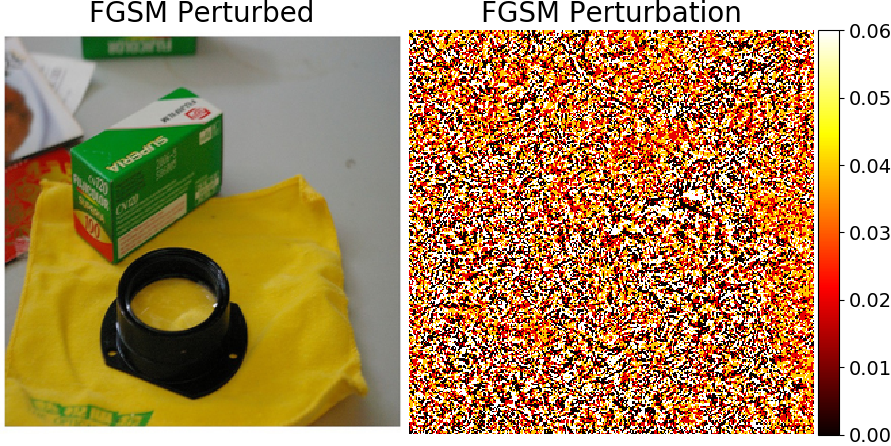}}}
\caption{DenseNet121 adversarial examples with the heat map of pixel intensities corresponding to the attack perturbations for the original image with the label {\bf loupe}.}
\end{figure}

\begin{figure}[h!]
\centering
\subfloat[bucket]{{\includegraphics[scale=0.32]{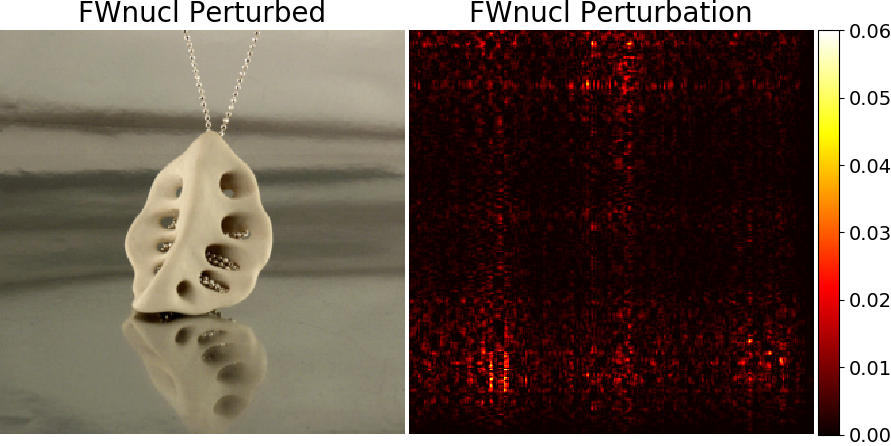}}}\hfill
\subfloat[carton]{{\includegraphics[scale=0.32]{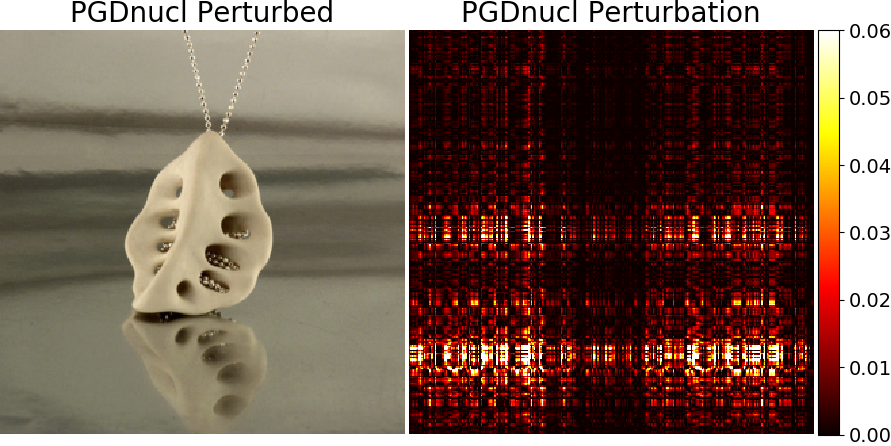}}}\\
\subfloat[horizontal bar]{{\includegraphics[scale=0.32]{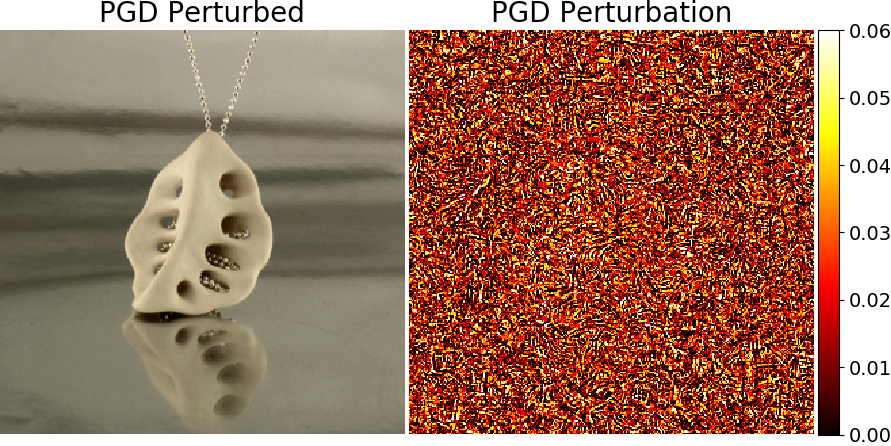}}}\hfill
\subfloat[iron]{{\includegraphics[scale=0.32]{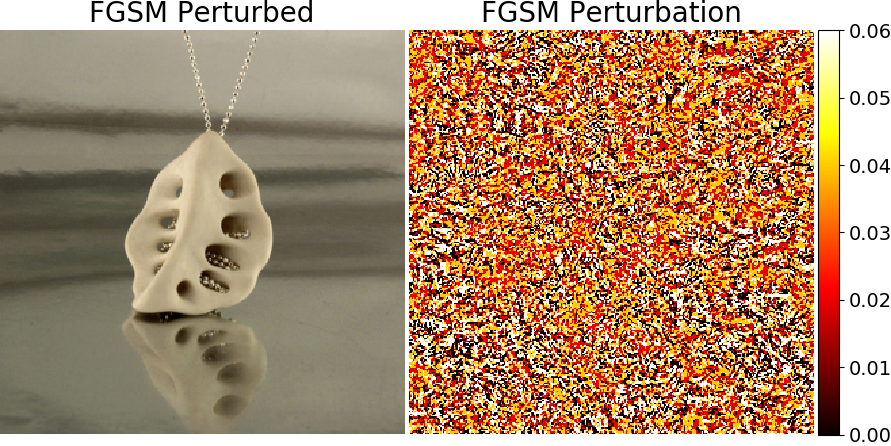}}}
\caption{DenseNet121 adversarial examples with the heat map of pixel intensities corresponding to the attack perturbations for the original image with the label {\bf necklace}.}
\end{figure}

\begin{figure}[h!]
\centering
\subfloat[knot]{{\includegraphics[scale=0.32]{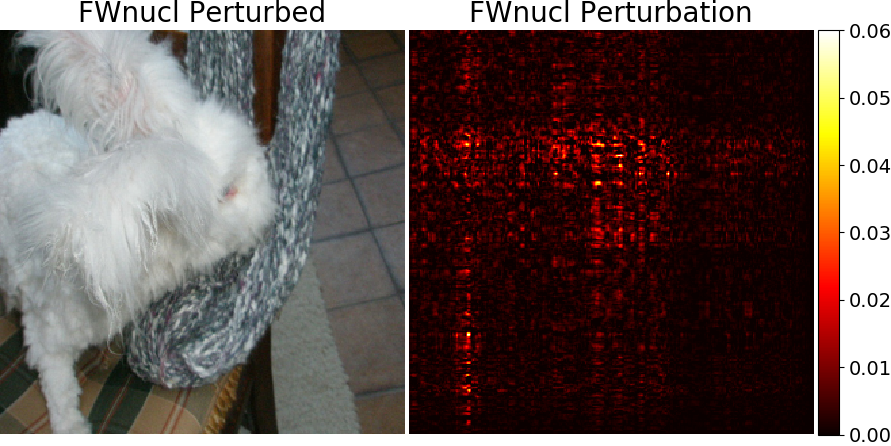}}}\hfill
\subfloat[wool]{{\includegraphics[scale=0.32]{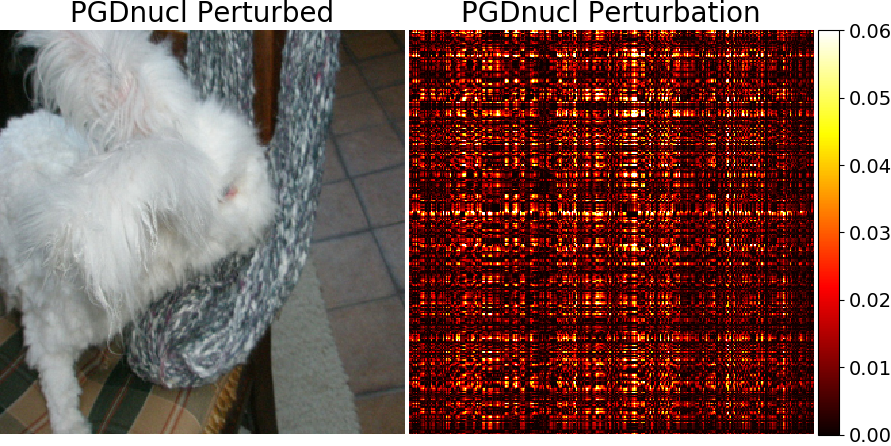}}}\\
\subfloat[mitten]{{\includegraphics[scale=0.32]{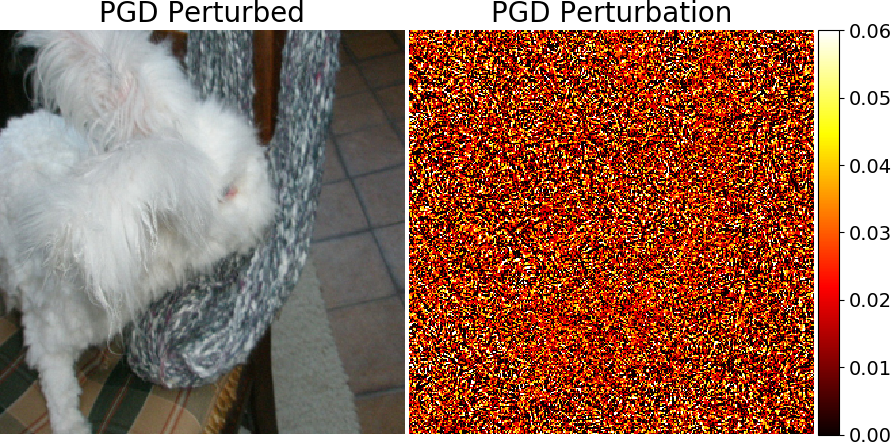}}}\hfill
\subfloat[mitten]{{\includegraphics[scale=0.32]{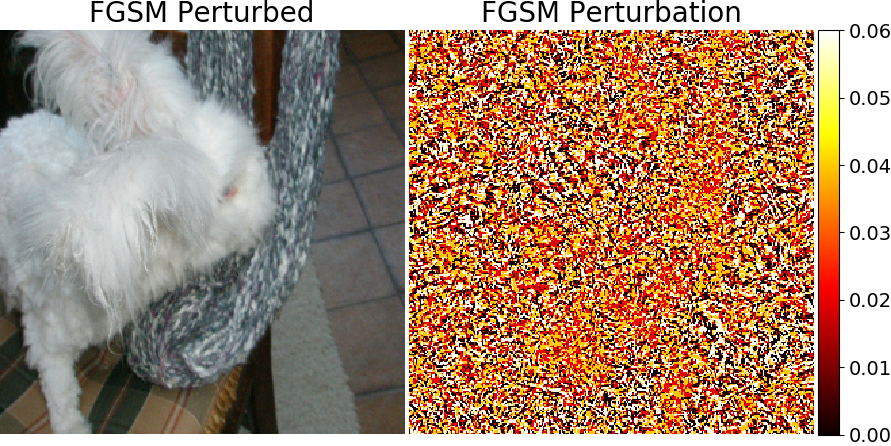}}}
\caption{DenseNet121 adversarial examples with the heat map of pixel intensities corresponding to the attack perturbations for the original image with the label {\bf Angora}.}
\end{figure}

\begin{figure}[h!]
\centering
\subfloat[plate rack]{{\includegraphics[scale=0.32]{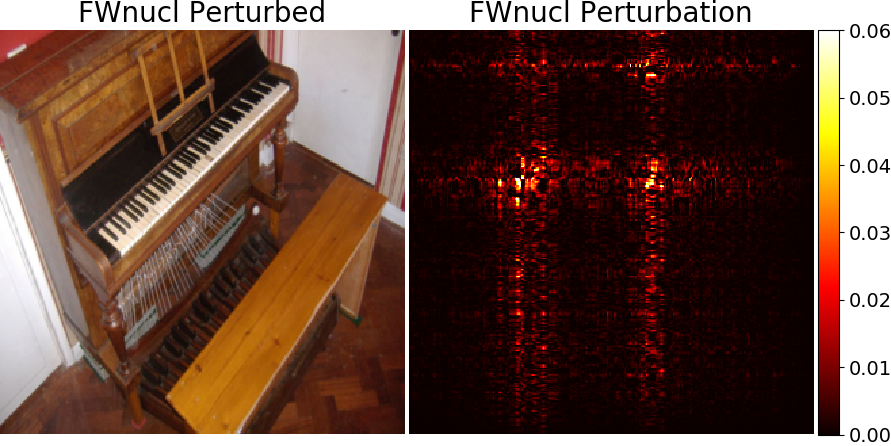}}}\hfill
\subfloat[upright]{{\includegraphics[scale=0.32]{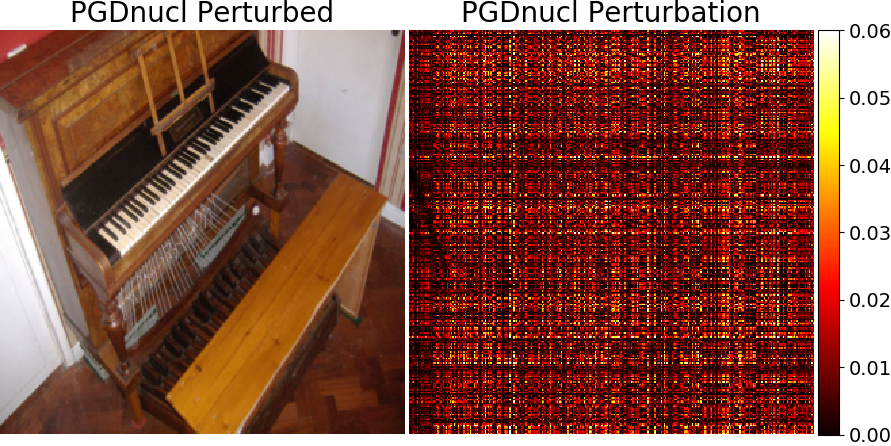}}}\\
\subfloat[abacus]{{\includegraphics[scale=0.32]{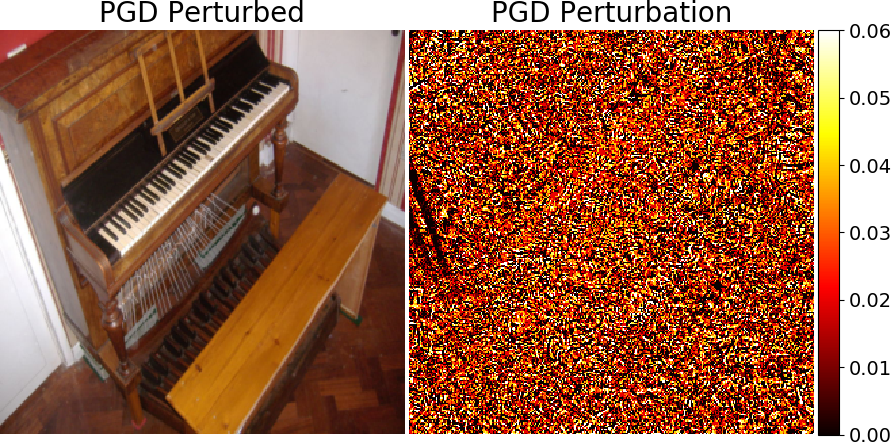}}}\hfill
\subfloat[plate rack]{{\includegraphics[scale=0.32]{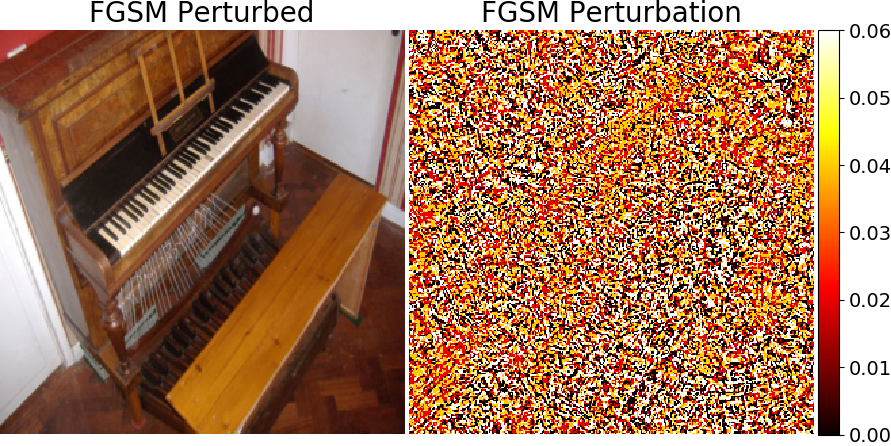}}}\caption{DenseNet121 adversarial examples with the heat map of pixel intensities corresponding to the attack perturbations for the original image with the label {\bf upright}.}
\end{figure}

\begin{figure}[h!]
\centering
\subfloat[Staffordshire bullterrier]{{\includegraphics[scale=0.32]{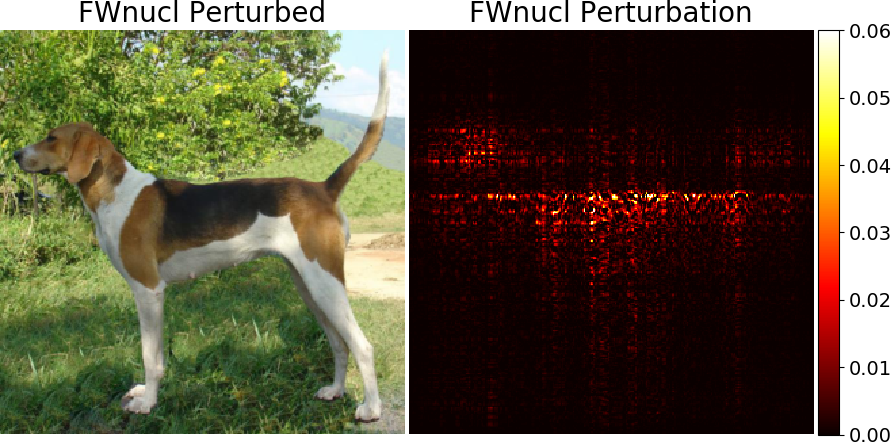}}}\hfill
\subfloat[English foxhound]{{\includegraphics[scale=0.32]{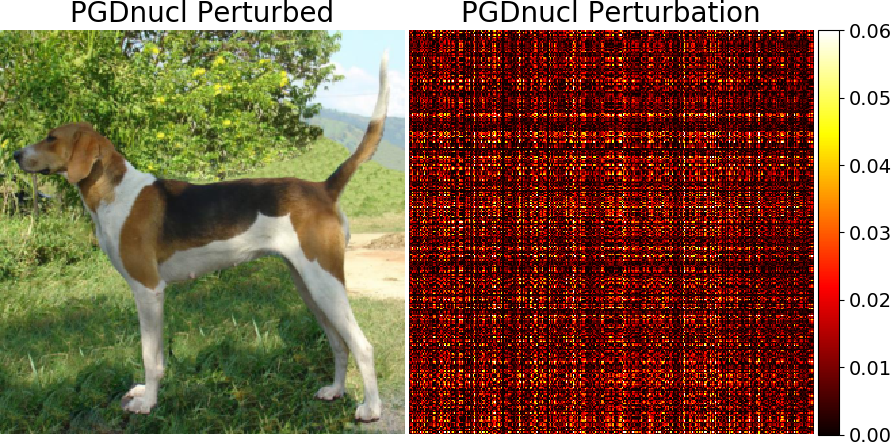}}}\\
\subfloat[beagle]{{\includegraphics[scale=0.32]{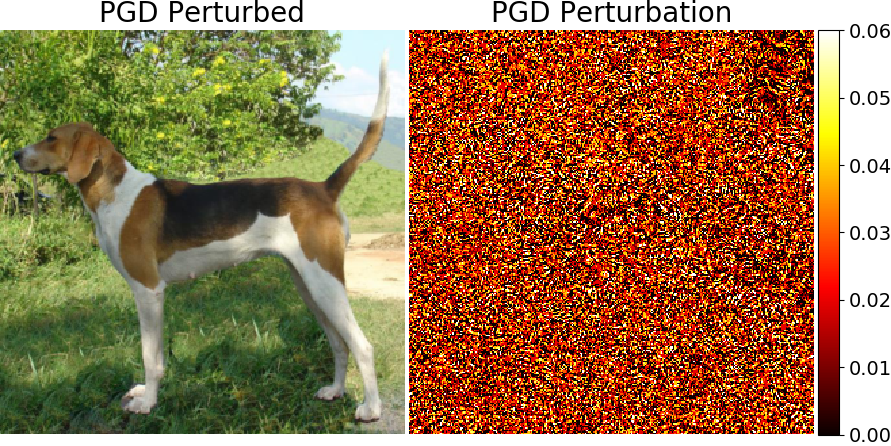}}}\hfill
\subfloat[beagle]{{\includegraphics[scale=0.32]{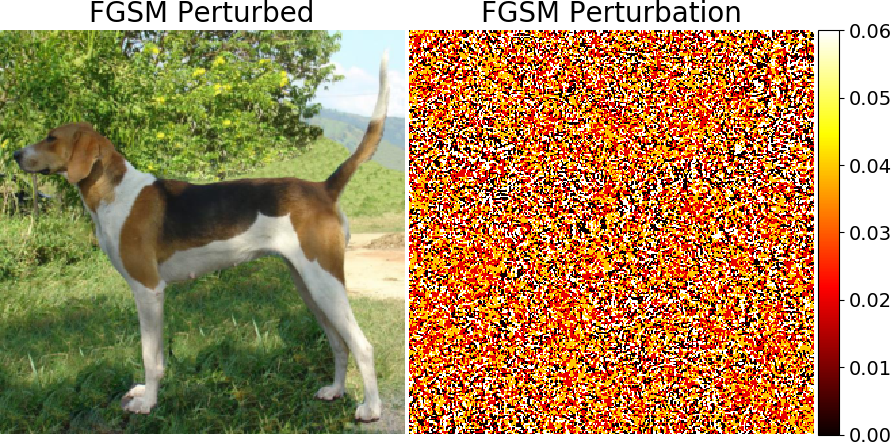}}}
\caption{DenseNet121 adversarial examples with the heat map of pixel intensities corresponding to the attack perturbations for the original image with the label {\bf English foxhound}.}
\end{figure}

\end{document}